\documentclass[journal]{IEEEtran}
\IEEEoverridecommandlockouts                              

\usepackage{url}
\usepackage{amsfonts}
\usepackage{graphicx}
\usepackage{subfig}
\usepackage{rotating}
\graphicspath{{figures/}}
\usepackage{multirow}
\usepackage{hhline}
\usepackage{amsmath}
\usepackage{adjustbox}
\usepackage{caption}
\usepackage{fancyhdr}
\usepackage{rotating}
\usepackage{hyperref}
 \usepackage[table,xcdraw]{xcolor}

\usepackage{calrsfs}
\DeclareMathAlphabet{\pazocal}{OMS}{zplm}{m}{n}

\newcommand{\RNum}[1]{\lowercase\expandafter{\romannumeral #1\relax}}
\usepackage{array}
\newcolumntype{C}[1]{>{\centering\arraybackslash}p{#1}}
\renewcommand{\arraystretch}{1.3}
\usepackage{multirow}
\makeatletter
\newcommand{\removelatexerror}{\let\@latex@error\@gobble}
\makeatother
\usepackage[linesnumbered,ruled,vlined]{algorithm2e}
\SetKwInput{KwInput}{Input}                
\SetKwInput{KwOutput}{Output}              
\usepackage{comment}
\usepackage{lipsum}
\usepackage{multicol}

\usepackage{booktabs}
\newcommand{\ra}[1]{\renewcommand{\arraystretch}{#1}}

\ifCLASSINFOpdf
\else
\fi

\begin{document}
%

\title{Slip Detection and Grip Force Control for Event-based Finger Vision}%
\title{Neuromorphic Vision Based Incipient Slip Detection and Correction for Effective Robotic manipulation }%
\title{Neuromorphic Vision-Based Slip Detection and Supression for Robotic grasping and manipulation }%
\title{Neuromorphic Vision-Based Slip Detection and Correction Method for Robotic Manipulators }%
\title{Neuromorphic Vision based Slip Detection and Suppression in Robotic Grasping and Manipulation }%
\title{Neuromorphic Event-Based Slip Detection and Suppression in Robotic Grasping and Manipulation }%
\author{Rajkumar~Muthusamy$^{1}$\href{https://orcid.org/0000-0002-5372-0154}{\includegraphics[scale=0.75]{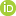}},~Xiaoqian Huang$^{1}$\href{https://orcid.org/0000-0002-2782-9068}{\includegraphics[scale=0.75]{figures/orcid.png}}, ~Yahya~Zweiri$^{1,2}$\href{https://orcid.org/0000-0003-4331-7254}{\includegraphics[scale=0.75]{figures/orcid.png}}, Lakmal Seneviratne$^{1}$\href{https://orcid.org/0000-0001-6405-8402}{\includegraphics[scale=0.75]{figures/orcid.png}} and ~Dongming~Gan$^{3}$\href{https://orcid.org/0000-0001-5327-1902}{\includegraphics[scale=0.75]{figures/orcid.png}} 
\thanks{$^{1}$  Khalifa University Center for Autonomous Robotic Systems (KUCARS), Khalifa University of Science and Technology, Abu Dhabi, UAE. Email: {\tt\small \{rajkumar.muthusamy@ku.ac.ae\}} }
\thanks{$^{2}$ Faculty of Science, Engineering and computing, Kingston University, London SW15 3DW, UK. }
\thanks{$^{3}$ School of Engineering Technology, Purdue University, West Lafayette, IN 47907, USA. }
}

\markboth{Journal of \LaTeX\ Class Files,~Vol.~14, No.~8, August~2015}%
{Shell \MakeLowercase{\textit{et al.}}: Bare Demo of IEEEtran.cls for IEEE Journals}

\maketitle

\begin{abstract}

Slip detection is essential for robots to make  robust grasping and fine manipulation. In this paper, a novel  dynamic vision-based finger system for slip detection and suppression is proposed. We also present a baseline and feature based approach to detect object slips under illumination and vibration uncertainty. A threshold method is devised to autonomously sample noise in real-time to improve slip detection. Moreover, a fuzzy based suppression strategy using incipient slip feedback is proposed for regulating the grip force. A comprehensive experimental study of our proposed approaches under uncertainty and  system for high-performance precision manipulation are presented. We also propose a slip metric to evaluate such performance quantitatively. Results indicate that the system can effectively detect incipient slip events at a sampling rate of 2kHz ($\Delta t = 500\mu s$)  and suppress them before a gross slip occurs. The event-based approach holds promises to high precision manipulation task requirement in industrial manufacturing and household services. 
\end{abstract}


%
\IEEEpeerreviewmaketitle

%
%
%
%
%
%

\section{Introduction}

\begin{figure}[!th]
  \centering
   \subfloat[]{\label{fig:sf1}
      \includegraphics[width=0.23\textwidth, height=0.23\textwidth]{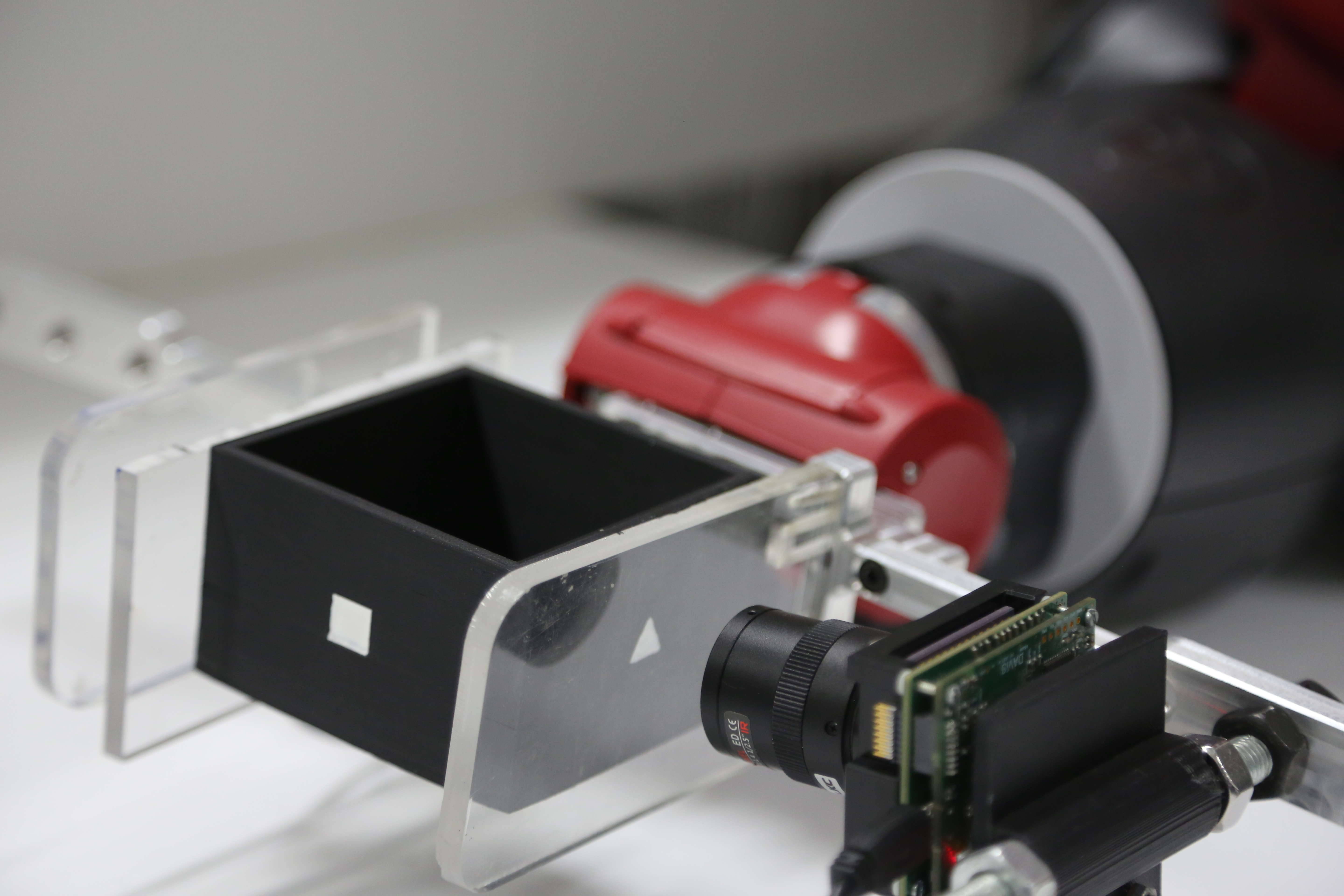}}  
\hfill
        \subfloat[]{\label{fig:sf1}
      \includegraphics[width=0.23\textwidth, height=0.23\textwidth]{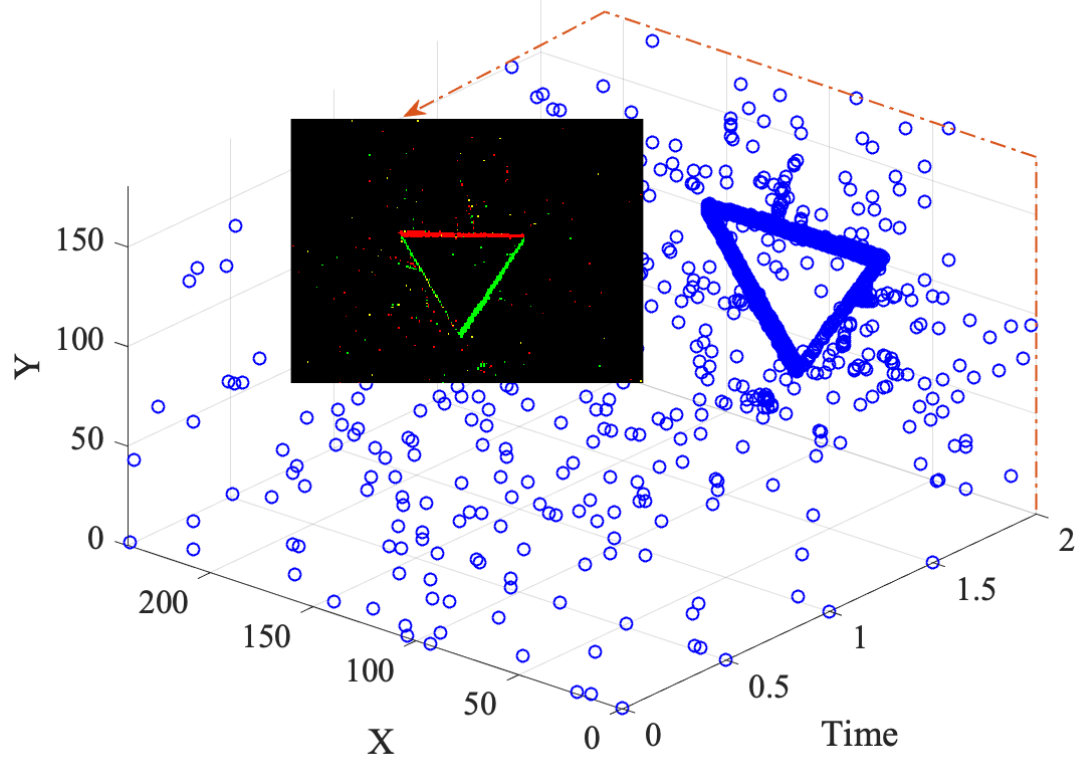}}\\
      \subfloat[]{\label{fig:sf1}
      \includegraphics[width=0.48\textwidth, height=0.3\textwidth]{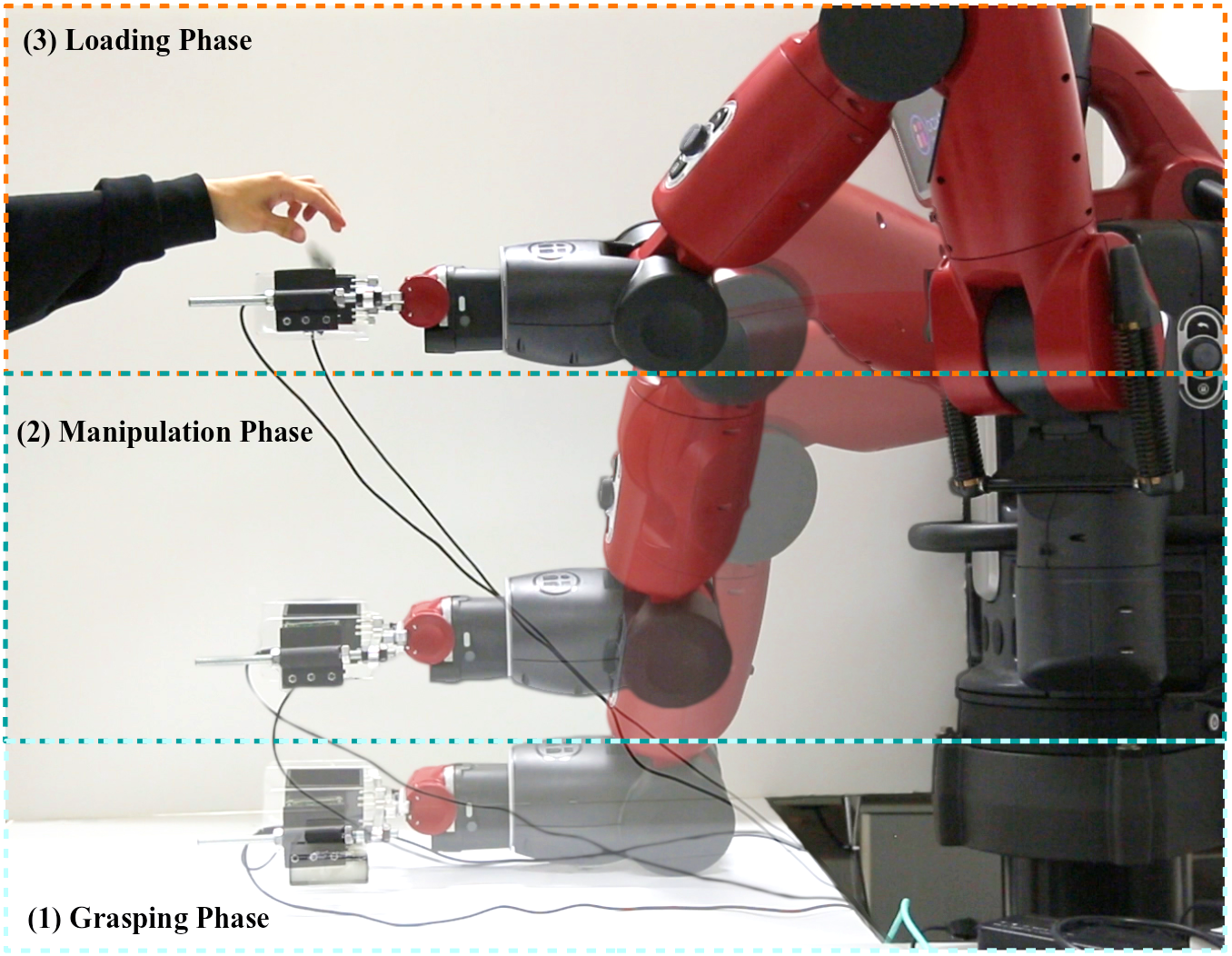}}
      \begin{tabular}{c|c}
     
      \end{tabular}
\caption{(a) Event-camera (DAVIS 240C) based finger prototype (b) Spatio-Temporal representation of a slip incident. Image shown at a particular temporal location corresponds to the projection of accumulated events over a time interval. (c) Slip incidents addressed within the three phases of robotic object manipulation.}
\label{first_pic}\vspace{-1.7em}
\end{figure}

With the emergence of Industry 4.0 \cite{aceto2019survey} and increased use of service robots in several application domains \cite{de2019grasping}, there is a growing demand for advanced perception capabilities in robotic systems to especially tackle the uncertainties occurring during physical tasks. Most of today's robots are equipped with parallel grippers or sophisticated hands which allows them to perform simple grasping to dexterous manipulation tasks \cite{birglen2018statistical} in both structured and unstructured environment. Slip incidents are common while performing such tasks under those settings. Slippage detection and suppression are key features for robotic grippers to achieve robust grasping and successful manipulation. Tactile and vision are the most important sensing modalities that endow robotic grippers with perception abilities to especially tackle slip incidents. Several types of sensors \cite{chen2018tactile} and methods have been addressed to detect and suppress such incidents. However,
the increased expectation of robots on high precision requirements of tasks \cite{li2019survey}, timely detection of transient changes in dynamic scenes and efficient acquisition and processing of sensory information enabling real-time response naturally attracts neuromorphic sensors \cite{vanarse2016review, liu2010neuromorphic}. Such event-based sensors emulate the perceptual power of biological systems. 

A recent technology and market report \cite{yole2019} predicts the neuromorphic sensing \cite{vanarse2016review, liu2010neuromorphic, indiveri2000neuromorphic} and computing technologies will make a great impact on automotive, mobile, medical, industrial and consumer sectors from 2024. At present, the development of neuromorphic vision sensors remains the primary focus of the neuromorphic sensing ecosystem. In this work, for the first time, a neuromorphic vision sensor 
that emulate the simplified neuro-biological model of a human eye retina is used 
for enhancing the physical (tactile) sensation of a robotic gripper at the finger level. In particular, we directly use the sensor with transparent finger material
 to detect slip with high temporal resolution and refer it as event-based finger vision that neither requires the object to be occluded nor any specialized or deforming skin between the object and sensor. Such settings offer cheap finger replacements, minimize wear and tear and increases slip detection accuracy and precision since the contact surface is not affected by the weight, material type, and geometry of the object. Fig. \ref{first_pic} (a) illustrates Baxter gripper with event-based finger vision prototype grasping a box carved with a primitive shape on each side.
  
Unlike conventional vision sensor which is frame-based and clock-driven, neuromorphic vision sensor is event-driven and has low latency, high temporal resolution and wide dynamic range. Moreover, the independent sensor pixels operate asynchronously and in continuous time responds to varying illumination. We exploit this inherent property of the sensor to achieve more effective, efficient and less resource-demanding detection of slip to tackle slip incidents in robotic object manipulation. In robotic applications, slip detection is considered with frequencies that are with in the range of 5-100 Hz. In this work, we present an event-based finger vision system and corresponding method for detecting incipient slip at a rate of 2kHz. The stream of events generated at the time of slippage is represented in a Spatio-temporal form in Fig. \ref{first_pic} (b). Image shown at a particular temporal location corresponds to the projection of accumulated events over a time interval.


 

In robotic grasping and manipulation, slip incidents may occur when (1) a grasp is executed with improper grasping strategy (2) lifting with insufficient force (3) the dynamic motion of the manipulator impacts the grasped object (4) the grasped object is subject to external disturbances such as addition of weight and placing back the object on the surface. Such slip incidents are accommodated within the three phases of robotic object manipulation demonstrated in Fig. \ref{first_pic} (c).   In \cite{ieng2014asynchronous}, the authors emphasized neuromorphic sensors are noisy and the difficulty in setting the ground truth measurement of noise in the event stream.

 Neuromorphic vision sensor suffers from increased sensitivity to varying illumination and small vibration caused by robot motion especially in unstructured environment. In particular, compliant robot performing manipulation tasks causes more noise in the event data due to continuous vibrations and augmented illumination uncertainty from robot compliance. We propose a feature-based approach to tackle these noise events and robustly detect object slips. Sensing modalities such as vision and tactile that involve neural processing found to extract high level geometric features of a stimulus at a early stage of processing pathways \cite{gollisch2010eye,bensmaia2008representation,yau2009analogous} such as edge orientation at the level of first order in both tactile neurons \cite{pruszynski2014edge} and neurons in the retina \cite{venkataramani2010orientation}. Inspired by the biological processing methods, we extract the corner and edge features from the temporal evolving events during a slip motion. These features assists by distinguishing real slippage and noise induced spurious slip events which is sampled at a high frequency.


 Another issue related to motion based methods for slip sensing is how to assign a threshold value to distinguish actual slips and noise. Mostly it is done empirically or by offline training and calibration. However, these methods are only suitable for static environments. For our event-based finger vision based gripper, we propose a method that autonomously sample noise thresholds online to perform a given manipulation task. Moreover, finding an appropriate grip force to tackle the varying slips is challenging. Fuzzy logic control \cite{iancu2012mamdani} incorporates high-level humanlike IF–THEN rule thinking and reasoning. Such model free control has been successfully applied to a wide variety of practical problems. 
 To suppress slip, a fuzzy logic based control scheme  is devised to regulate the grip force using incipient slip feedback. 

In the following, we systematically address the areas of neuromorphic sensing, conventional sensing, slip detection and suppression in more details. 
 



\subsection{Neuromorphic Sensing}

Sense of touch \cite{dahiya2009tactile} and vision is the most important sensory modalities that allows humans to maintain a controlled grip. Neuromorphic sensors offer viable solution to emulate the processing of sensor signals from such biological modalities. 
In general, a neuromorphic sensor mimics neuro-biological architecture \cite{maher1989implementing}  rather than emulating a complete sensory system.  Such sensors encode sensory information into time-series of spikes which is asynchronous, sparse, rich in nature and use the temporal contrast of spikes to encode a wide range of information based on the application requirements.  Moreover, they minimize the amount of redundant data transmission by capturing transient changes in the dynamic scene.

Human hand is the most specialized part of the body that provides accurate tactile feedback \cite{vallbo1984properties}. 
Detection of incipient slip is one key functionality of the tactile sensing modalities which enables human to perform robust grasping and dexterous manipulation.
In particular, human hand posses four functionally distinct tactile receptors distributed and overlapping in the uneven skin surface. These receptors are classified into fast adapting (FA-I, FA-II) and slow adapting (SA-I, SA-II) which responds to skin deformation and vibrations at a frequency up to 400 Hz \cite{westling1987responses}. During tactile activity, the signal pattern from different receptors are diverse for different tasks and their combination increases the level of pattern complexity. Difficulties in 
obtaining a clear model for such complex biological system is one of the primary reason for the limited progress in artificial tactile sensing and development of neuromorphic tactile sensor. Alternatively, neuromorphic approach is used to transform tactile signals to biological relevant representation (spike events).  Recently, drawing inspiration from the behaviour of mechanoreceptors (e.g. FA-I and SA-I afferents), \cite{romano2011human} demonstrated the feasibility of a tactile-event-driven model for grasp control, \cite{nakagawa2019bio} developed a slip detection and suppression strategy for robotic hands.



Vision is one of the most important sensing modalities heavily used by humans for perception. In fact, the retina is the extensively studied human neuro-biological system which remains a prominent example for the model, design and implementation of neuromorphic sensors \cite{jung2011biohybrid}. The retina is a thin layer of tissue lined at the back of the eye ball which is mainly composed of photoreceptors, bipolaar cells, and ganglion cells \cite{gollisch2009throwing}. The complex network between photoreceptors to ganglion cells in the retina converts light in to electric impulses (spikes) that is relayed to brain via an optic nerve. The spikes produced by the ganglion cells carry visual information which is encoded on the basis of spike rate, spatial-temporal relation, temporal contrast or any of these combinations. In particular, the X and Y ganglion retina cells and their retina-brain pathways gives insights on 'what' and 'where' information from the biological vision system. The 'where' system is sensitive to changes and motion and oriented towards detection with high temporal resolution. The 'what' system transports detailed spatial, texture, pattern and color information.

 
Conventional frame-based image sensors are focused on implementing the 'what' system by which they neglect the dynamic information in the visual scene. Recently, Dynamic vision sensor (DVS) \cite{lichtsteiner2008128} was mainly developed to realize the 'where' system. The DVS sensor constitutes a simplified three layer model of the human retina that operates in continuous time by responding to brightness changes in the scene. Each individual pixel in the sensor array works autonomously and respond to temporal contrast by generating asynchronous spiked events. DVS established a benchmark in neuromorphic vision sensing and was used in robotics applications involving high-speed motion detection and object tracking. Further exploiting biological vision system, an Asynchronous Time Based Image Sensor (ATIS) \cite{posch2010qvga} was developed which is a combination of  'where' and 'what' systems that contains event-based change detector to output a stream of timed spikes and pulse width modulation based exposure measurement units to encode absolute intensity into the timed spikes.  The DAVIS \cite{berner2013240, brandli2014240} is a combination of an asynchronous 'where' system and a synchronous 'what' system. It outputs event-based frames through the synchronous active-pixel sensor and simultaneously outputs events through the asynchronous DVS sensor. Neuromorphic vision sensors are recently commercialized and many companies such as IniVation, Prophesee, Sony, Insightness and celepixel are in the process of industrial-grade mass production \cite{Anne2019}. In our work, we exploit only the dynamic vision sensing part of DAVIS for incipient slip detection in robotic manipulation and our proposed approaches directly process object motion changes in real-time.

\subsection{Slip Detection Via Conventional Sensing}

In robotic grasping and manipulation, slip incident may occur when a grasp is executed with improper grasping strategy or insufficient force or grasped object is subject to external disturbances.  Incipient slip and gross slip are two main states and contiguous phenomenon of slippage where the incipient slip take place prior to gross slip. Incipient slip refers to a state at which the object start to loose its boundaries under grasped condition. If such state of slippage is uncontrolled, then a further displacement of object occurs which then leads to a state of gross slip. 


 In robotic grasping, grasp planning \cite{Bohg2014} is mainly conducted in simulation environments since it involves exhaustive search and evaluation of grasp hypothesis for a given object and robotic hand. Such controlled way of planning avoids the necessity of full scale experiment with real hardware which is time consuming and costly.
Apart from that, execution of the determined grasp in pose and contact level is not practical.
Further, a manipulation task with a grasped object is subject to internal and external disturbances due to the dynamic motion involved. Even these task-oriented disturbances \cite{LiSastry1988} are modeled to a certain extent based on demonstrated experiments \cite{LinSun2015} and heuristics then used in grasp planning. However, the contact models \cite{salisbury1983kinematic}, criteria \cite{Ferrari1992} and quality measures \cite{Roa2015} used for determining a grasp is only a close approximation which cannot be ideally devised for practical situations. Therefore a sensory feedback is required to adjust the grasping force in all stages of robotic object manipulation. 
Slippage can be a rich sensory feedback for robotic hands to tackle object stability issues during grasping and object manipulation.

Tactile sensing still remains a key element in the process of robotic manipulation. Robotic grippers and hands are increasingly equipped with different types of tactile sensors. Based on the working principles, tactile sensing is mainly achieved by detecting  object motions directly or indirectly. In the following, we address the tactile sensing methods that use indirect ways for slip detection. (1) observing the ratio of the measured tangential force to the measured normal force at the contact point by using F/T sensor \cite{melchiorri2000slip}. (2) Measuring changes in shear force using center-of-pressure (COP) tactile sensors  \cite{gunji2007grasping, mizoguchi2010development}. (3) measuring and analysing the vibration of the shearing force caused by relative motions by sensors such as acoustic resonant sensors \cite{shinoda2000instantaneous}, thick-film piezoelectric sensors \cite{cotton2007novel}, piezoresistive sensors \cite{dao2011development}, and optoelectric tactile sensors \cite{dubey2006dynamic}, etc. (4) physically observing the slip displacement of the object from robotic hands using various sensors such as optical sensors \cite{roberts2011slip}, accelerometers \cite{tremblay1993estimating}, array tactile sensors \cite{yussof2010sensorization}. Tactile sensing hardware and technologies are still underdeveloped when compared to other perception modalities such as vision. The slow phase of development is due to the realization of inherent complexity in the sense of touch in human hands. Another important reason is conventional tactile sensors requires direct contact with the object that is subject to wear and risk of sensor saturation and damage which makes conventional tactile sensing a non attractive solution for industrial applications.

 The idea of using frame-based image sensors for tactile sensing is not new which allows detecting of object motion. Detecting the internal reflections via optic based vision sensing \cite{yussof2010sensorization}, marker-based: markers are placed on the sensor surface and their displacement was measured using image processing techniques \cite{ito2010shape, assaf2014seeing, hristu2000performance} and registration of object through markerless localization \cite{li2014localization}. In most of the works, vision sensors are placed underneath the skin surface to detect the motion of markers which somehow limits the ability of the vision sensor in distinguishing whether the change of contacts are from grasped object or external disturbances. More recently, Yamaguchi et al. \cite{yamaguchi2017grasp} proposed a vision based tactile sensor \cite{yamaguchi2016combining} to detect slip and proposed a grasp adaptation controller that modifies the grasp to avoid slip.

\subsection{Slip Prevention and Suppression Methods}
Slip prevention methods can be mainly classified under pre-grasping and post grasping phases. In the Pre grasping phase, the grasping strategy focuses on where and how to grasp a known object such that the executed grasp is robust against disturbances. This provides an appropriate grasp to prevent the occurrence of slip. Several model based approaches to analyse the grasp properties and quality measures to quantify grasp quality have been proposed in \cite{Roa2015}. In \cite{RV14} non-task specific and task specific metric \cite{RV2016} are used to quantify the disturbance rejection property of force closure grasps \cite{Nguyen1988}.
In post-grasping phase, several approaches focus on regulating the grip force to tackle the slip incidents that occur during object manipulation tasks.
For a detected slip, (1) the controller increments force in small percentage until the slip stops \cite{romano2011human}. This may cause the object to reorient or  squeeze (2) the controller increases the desired grip force proportional to the magnitude of the slip event \cite{takahashi2008adaptive}. Such controller requires additional sensor to sense object motion and algorithms for processing such information. In our control scheme, we compute the magnitude of the incipient slip detected by the event-based finger vision and use a mamdani-type fuzzy logic controller to regulate the normal component of grip force until the slip stops. Our fuzzy controller only requires statistical data of slip magnitudes from multiple repeated experiments with varying loads in order to set the min and max values for the fuzzy sets such that the grip force is determined based on the rules.




%
%
%

\subsection{Contributions}
A rich survey on event-based vision is available in \cite{gallego2019event} where several areas relating to robotic applications such as object recognition and tracking, pose tracking and Simultaneous Localization and Mapping (SLAM) are reviewed. Slip detection is a challenging problem in robotic grasping and manipulation. In this paper, we present an approach of detecting slip with event-based vision sensor. In particular, we developed an event-by-event approach where the stream of events that occurs in a microsecond range are directly processed to detect object slips.
Only few recent works addressed dynamic vision sensing in robotic grasping. In \cite{rigi2018novel}, an event-based frame approach to detect incipient slip between a $10 ms$ to $44.1 ms$ sampling rate was presented. At the contact level, silicon material was used and the event distribution was analysed on the reconstructed frames for slip detection. In \cite{naeini2019novel}, machine learning methods were used to estimate contact force and classify materials for a grasp. The authors in \cite{ward2020neurotac} presented a NeuroTac sensor and corresponding encoding methods for texture classification task. They found timing based coding method gave highest accuracy over both artificial and
natural textures. Our proposed slip detection approaches are based on temporal coding.

For the first time, an event camera based approach is developed to detect passively incipient slip and gross slip at a $500 \mu s$ sampling rate which could make it a good candidate for industrial and collaborative robots applications.
Moreover, an intelligent slip suppression strategy using the incipient slip feedback to adjust the grasping force is devised.

In the following,  the primary contributions of this paper are summarized.

%
%
%
%
%
%
%
%
%

\begin{enumerate}
\item We present an event-based finger vision system and a method to detect and suppress slip in a timely manner using event data. In particular, the method initiates three stages of process to conduct a task and calibrates the slip detection algorithm autonomously for online operation.



\item We propose and comprehensively study two event-based slip detection approaches, a baseline and a feature based for robust detection of object slips under illumination and vibration uncertainty.

\item We design a mamdani-type fuzzy logic to adjust the grasping force of the robotic gripper using event-based incipient slip feedback.



\item We demonstrate experimentally neuromorphic vision based slip detection and suppression in the phases of robotic object manipulation. Especially, for object slips caused by insufficient grasp force while lifting, speedy manipulation operation, loading under grasped condition and surface contact while placing. We propose a slip metric to evaluate the performance of the overall task.


\end{enumerate}

\section{Event-Based Slip Detection and Suppression Method}

A primary goal of a robotic grasp is to immobilize an object to allow precise manipulation. Form and force closure are the well known conditions to maintain object immobility. Form closure considers grasp geometry that kinematically constraint an object whereas force closure considers forces applied by the frictional fingers to withstand external wrenches applied on the object.

Force closure is a minimal condition that uses static friction (Coulomb friction model) to prevent slipping between two bodies. To avoid slippage, a contact force $\boldsymbol{\vec f}$ at a point $j$ must satisfy the frictional constraint
\begin{equation} \label{fric_const}
FC_{j}= \{ \boldsymbol{\vec f}_{j} \in \Re^{3} | \sqrt{f_{o_{j}}^{2} + f_{t_{j}}^{2}} \,  \leq \, \mu f_{n_{j}} \} \end{equation}

 where $\mu $ is the empirically determined coefficient of friction that bounds the tangential components  $f_{o_{j}},f_{t_{j}}$ with respect to the applied normal component $f_{n_{j}}$ at the contact point. In short, all admissible forces by a contact normal are constrained to the friction cone $FC_{j}$. However, there are infinite possibilities of contact force values that can be applied while grasping. In most cases, a minimal force is applied by the grasp to avoid damages to both object and robotic gripper. When the grasping forces are not adequate, the friction coefficient decreases and causes slip. Sensory-based information can be used to effectively tackle slip incidents and enforce force closure under uncertain conditions.

  At the time of disturbance, errors are caused in the placement of contacts, object pose and finger force which lead to slippage. Thus, the robotic gripper needs sensory information to effectively detect slip and regulate grasp forces to compensate the disturbances in a way to maintain object stability. Detection of incipient slip is crucial for robotic gripper/hand to adjust the grasping force and provide a stable grasp.

\subsection{Event-based finger vision System }
\begin{figure}[th!]
\center
 \subfloat{\label{fig:sf1}
      \includegraphics[width=1 \linewidth]{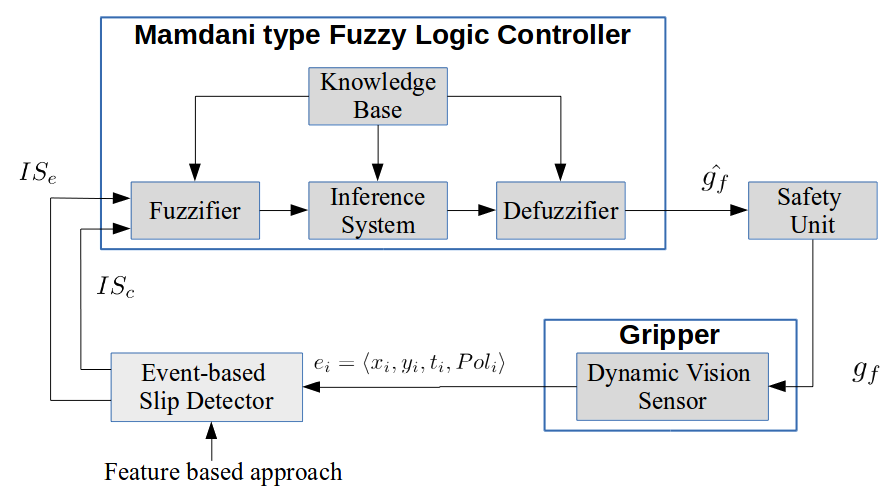}} 
      \hfill
\caption{Event based finger vision system diagram}
\label{fig2}
\end{figure}
An event-based finger vision system for slip detection and suppression is presented in Fig. \ref{fig2}. The gripper with event-camera based finger prototype takes the visual changes and outputs stream of events, briefly explained in Sec. \ref{DVS}. Our feature based slip detector classifies each event as corner, edge and flat points in real-time and evaluate whether it signals slip, detailed in Sec. \ref{e-Harris} \& \ref{slip_detection}. Then, the incipient slip which is the first instance of the temporal window from the detected slip is used in the mamdani type fuzzy controller detailed in Sec. \ref{slip suppression}. In particular, the number of edge and corner features detected in the incipient slip is used as inputs to determine an appropriate grip force. Moreover, the following safety unit regulates the grip force based on the magnitude of incipient slips. Then the desired grip force is sent to the gripper for actual slip suppression. 
\subsection{ Dynamic Vision Sensing } \label{DVS}

Dynamic vision sensor \cite{lichtsteiner2008128} has array of pixels that respond independently and asynchronously to logarithmic brightness $(L= log(I))$ changes in a scene. The illumination $(L (t))$ sensed  at the photorecepton of individual pixels is encoded in the form of temporal variance in the event based camera. More precisely, an event is generated at a pixel $(x,y)$ at time $t$ when the absolute difference of log intensity value reaches a temporal contrast threshold $ C^{\pm}$
\begin{equation}
\begin{split}
\bigtriangleup L(x,y,t) \: = \:
\mid L(x,y,t) - L (x,y,t- \bigtriangleup t) \mid
\\= Pol *\: C  \:\Bigg\vert \: C>0, Pol \in \{+1,-1 \}
 \label{eq1}
\end{split}
\end{equation}
where  $\bigtriangleup t$ is the arbitrary period of time elapsed since the last event at the same pixel and $Pol$ denotes events polarity with sign indicating the brightness increase and decrease. A threshold ranging between 15\% to 50\% of illumination change is set internally in the event based camera through electronic biases. In our case, we use DAVIS 240C dynamic vision camera which has a spatial resolution of (240 $\times$ 180 pixels) and dynamic range of 120 dB.

Event cameras represents visual information in terms of time with respect to a spatial reference in the camera-pixel arrays. Pixels in the dynamic vision sensor respond independently and asynchronously to logarithmic brightness changes in the scene. For a relative motion, a stream of events with a microsecond ($\mu s$) temporal resolution and latency is generated, where an event $e =\langle x,y,t^{f}, Pol \rangle$ is a compactly represented tuple in a spatio-temporal form. However, analysing a single latest event does not give much information in operational level and exploring all past events is not scalable.  Thus we opted the surface of active events (SAE) \cite{benosman2013event} for performing operations on the evolving temporal data in camera pixel-space. 
 The surface represents the timestamp of a latest event at each pixel from the event stream. For each upcoming event, the function $\Sigma_{e}: \mathbb{N}^{2} \mapsto \mathbb{R}$ takes the pixel position of a triggered event and assign to its timestamp:

\begin{equation}
\Sigma_{e} : (x,y) \mapsto t^{f} \: | \: (x,y) \in \mathbb{R} \times \mathbb{R}
 \label{eq1}
\end{equation}

Feature detectors reduce the event stream to a highly informative stream of events. It is a filter acting on the SAE that reduces the amount of data and computation cost for further high level processing such as slip detection. The feature detection methods process the stream of events in two ways: event-by-event and event-based frame. The first one directly operates on the asynchronous event stream \cite{clady2015asynchronous,vasco2016fast,mueggler2017fast,alzugaray2018asynchronous} whereas the second synthesize an artificial frame based on a fixed temporal window \cite{zhu2017event} or fixed number of events \cite{vidal2018ultimate}. In our slip detection approach, we consider event-by-event approach for detecting highly informative features.

\subsection{ Spatially Adaptive e-Harris } \label{e-Harris}


In conventional image processing, Harris detector is one of the most widely used technique that detects features such as corner, edge and flat points based on intensity variation in a local pixel neighborhood. This feature detector is known for its efficiency, simplicity and invariance to scaling, rotation and illumination.  Unlike conventional camera that records large amount of redundant data in sequence of frames, the DVS records only the changes in the visual scene as stream of events characterized by the pixel positions and its timestamps and does not include intensity measures. Therefore the frame based Harris detector cannot be directly applied on the SAE. Event-based adaptation of Harris detector is proposed in \cite{vasco2016fast} and \cite{mueggler2017fast} where each upcoming event is directly processed. Their method binarizes the SAE by the newest $N$ events for the whole image plane or locally around the current event.
%

Algorithm \ref{algo1} summarizes the spatially adaptive e-Harris \cite{mueggler2017fast}. The e-Harris feature detector mainly relies on the analysis of the eigenvalues of the autocorrelation matrix. If the e-Harris score is large positive value, the event is classified as corner whereas a negative value is considered as edge. The rest of the value which is in-between is considered as flat points. In this work, the adapted e-Harris detector is used to detect edge and corner features in a event-by-event basis from locally perceived information that is independent of the scene and sensor size. Moreover, the algorithm parameters are modified. Selected corner and edge threshold of $C_{th} = 10$ and $E_{th} = -0.01$ buffer of latest events $N = 20$ and a patch of 9$\times$9 pixels gave the best performance over a wide variety of data-sets.
\begin{figure}[!th]
 \removelatexerror
\begin{algorithm}[H]
\DontPrintSemicolon

  \KwInput{Stream of events $e_{i} = \langle x_{i},y_{i},t^{f}_{i},Pol_{i}\rangle$}
   \KwOutput{e-Harris score $H_{s}$}
   Create an surface of active events $(\Sigma_{e})$ w.r.t. the pixel array of the DVS camera\\
    \For{ each $e_{i}$}{ 
        Create an  $L$ pixels wide spatial window (patch) centered around the pixel of the latest event.\\
        Binarize the local patch with N latest events (0 and 1 represents the event absence and presence)\\
        Compute gradient of the binary surface with Sobel operator.\\
        Compute the Harris matrix with Gaussian smoothing filter window.\\
        Compute Harris Score $H_{s}$\\
        Update SAE\\
        return $H_{s}$
   	}

\caption{Spatially Adaptive e-Harris} \label{algo1}
\end{algorithm} 
\end{figure}

\subsection{ Robust Slip Detection } \label{slip_detection}

%
%
%

\begin{figure}[!h]
 \removelatexerror
\begin{algorithm}[H]
\DontPrintSemicolon

  \KwInput{Stream of events $e_{i} = \langle x_{i},y_{i},t^{f}_{i},Pol_{i}\rangle$, e-Harris Threshold $fn_{th}$, $fp_{th}$, Sampling bias $S_{bias}$, Timestep $Xms$ (e.g.$ 500 \mu s$)}.
  \KwOutput{Detected slip measures: Feature based  $(s_{e},s_{c})$ and Baseline  $(s_{r})$.}


    \For{ each $e_{i}$}{ 
Compute e-Harris score (Algorithm 1).\\

Classify each event as edges and corners based on e-Harris score $H_{s}$ and heuristically set negative and positive threshold $fn_{th}$, $fp_{th}$.\\


Start counting raw $C_{raw}$ , and feature  $C_{edge} , C_{corner} $ events.\\

   	\If {$\Delta t > Xms$}
  {
  $C_{raw} = 0 , C_{edge} = 0 , C_{corner} = 0$. 
  }
  
\If {Grasp planned}
{
Initiate the noise sampling process.\\

No robot action and intervention of object in the visual scene.\\
Determine the noise thresholds by taking the highest event count from overall sampling interval. Raw ($th_{rmax}$) and feature ($th_{emax},th_{cmax}$) based noise thresholds are obtained.
%

}

\If {Grasp Execution}
{
Initiate the grasping process.\\
Move the robot to a pre-grasp pose determined from grasp planning.\\
Set a minimal grip force $(g_{f}^{min})$for the gripper which ranges from 0-100 percent.\\

Cage the object or Execute the grasp.\\
}

\If {Task execution}
  {
Initiate slip Monitoring.\\
Initiate robotic object manipulation that includes grasping, lifting, loading, lowering and dropping.\\
\If {Baseline Approach}
{
\If{$(C_{raw} >= Th_{rmax} + S_{bias} * Th_{rmax})$ }
{
$s_{r}= C_{raw}$.\\

   	}

\If {$C_{raw} < Th_{rmax} $}
{
$s_{r}= 0$.\\

   	}
   	}
   	
\If {Feature Based Approach}
{
\If{$(C_{edge} >= Th_{emax} + S_{bias} * Th_{emax}) \quad \mathbf{AND} \quad  (C_{corner} >= Th_{cmax} + S_{bias} * Th_{cmax})$}
{
$s_{e}= C_{edge}$, $s_{c}= C_{corner}$.\\

   	}

\If {$C_{edge} < Th_{emax} \quad \mathbf{AND} \quad C_{corner} < Th_{cmax} $}
{
$s_{e}= 0$, $s_{c}= 0$.\\

   	}
   	}   	

   	}

   	}

\caption{Event-based Slip Detection}
\end{algorithm} \label{Slip_Detection_algo}\vspace{-1em}
\end{figure}

We define slippage for our proposed event based approaches in the following way: A gross slip ($s_{*}$) is the number of events accumulated from the continuous translation and rotational motion of the object at a desired sampling rate. An incipient slip ($is_{*}$) refers to the first instant of accumulated events at which the object start to loose its boundaries. 

 \textbf{Baseline Approach:} The stream of events from the dynamic vision sensor is directly processed by this approach to detect slip incidents. We consider a continuous time function  $e(t)$ which turns the triggered events to a sequence of spikes expressed as
 \begin{equation}
e(t)= \delta (t-t^{f})
\end{equation}
 where $\delta (t-t^{f})$ is a unit impulse function and $t^{f}$ is the time at which the firing of an event $e =\langle x,y,t^{f}, Pol \rangle$ occurs. We consider a temporal window of width $\Delta t$ rolling over the spiking train in timesteps of $500\mu s$. The step size is kept equal to the sample time. Within the temporal window ($\Delta t$), the total number of spikes are recorded before proceeding to the next timestep. To detect slip incidents, the baseline approach take the sum of the spike count from each sampling period and checks whether it crosses a threshold. The baseline approach can be expressed as 

\begin{equation}
BA(t)=
    \begin{cases}
      1, & \text{if}\ \sum_{t=t}^{t+\Delta t} e(t) > Th_{rmax} \\
      0, & \text{otherwise}
    \end{cases}
 \label{ba}
\end{equation}

where $Th_{rmax}$ is the noise threshold which is determined by taking the maximum spike count from the array of sliding sums over a time period. Such sampling procedure is performed while grasp planning. The approach rejects slip hypothesis when the spike events are insignificant and distinguish noise events from actual slip events. However, dynamic motion of the manipulator under varying light conditions and compliance in such manipulator causes more uncertainty in the event data. Thus, this approach may indicate noise events as actual slips in the fast detection process. Therefore we propose a second approach that is robust to such uncertainties and process highly informative event data.

\textbf{Feature based Approach:} On each upcoming event from the event stream, the adapted e-Harris in algorithm \ref{algo1} is used to detect highly informative feature events such as edges and corners. Let $F_{d}(e_{i})$ be the feature detector that classifies the events as corners and edges. Such featured events are labeled ($label=\{corner, edge\}$) and triggered at time $t^{f}_{label}$. Spikes triggered corresponding to a feature event is given as
 \begin{equation}
E_{label} (t)= \delta (t-t^{f}_{label}) 
\end{equation}
Similar to the previous approach, we slide the temporal window over the classified feature based spiking train. For each timestep, we accumulate the classified spikes separately and check whether it crosses corresponding feature based noise threshold. Thus, we detect the slip incidents by:
\begin{equation}
FBA(t)=
    \begin{cases}
      1, & \text{if}\ \sum_{t=t}^{t+\Delta t} \begin{cases} E_{edge}(t) > Th_{emax} \\  \wedge  \\  E_{corner}(t) > Th_{cmax}\end{cases} \\
      0, & \text{otherwise}
    \end{cases}
 \label{ba}
\end{equation}
 The approach robustly checks the  consistent variation of detected corners and edges by applying a simple AND logic operator and rejects slip hypothesis when any one of the feature is inconsistent in varying. 


 Several corner detection methods operating on the SAE and following the event-by-event approach was proposed. An intensity based Harris corner detector was adapted to an event level in \cite{vasco2016fast} and improved in \cite{mueggler2017fast} and referred as e-Harris. In the same work, they presented an efficient corner detector referred as e-Fast using comparison operators on the SAE. Recently, the ARC* \cite{alzugaray2018asynchronous} corner detector with enhanced detection repeatability and efficient than the other corner detector is presented. We mainly utilize the asynchronous event based corner detector (e-Harris) and adapt it according to our slip detection approach. Moreover, we study the above event based corner detectors performance in the context of slip detection in the experimental section. Any of the state of the art corner and feature detectors can be incorporated with the event-based slip detection algorithm.

Algorithm 2 summarizes the three stages of process and approaches for event-based slip detection. Grasp planning, grasp execution and task execution are primary step in robotic object manipulation. We integrate the event-based slip detection algorithm into these steps to autonomously calibrate them in real-time for online operation. Firstly, a robot with an event-based finger vision gripper plans a grasp for a known object in the scene. Simultaneously, the noise is sampled for a desired temporal window when there is no robot action or artificial intervention in the visual scene. The maximum value from the sampled intervals over time for the individual classified events are set as threshold and utilized in the corresponding proposed approaches. In particular,  The margin of threshold is increased by certain percentage (e.g. 10 \%) to reduce the sensitivity to noise and such bias is determined based on experimental noise analysis in section. \ref{noise_level_analysis}. Secondly, the  robot does the motion planning and reach the pre-grasp pose and executes the grasp with minimal grip force or does caging of the object. Finally, the object is monitored in task execution for any possible slips based on the proposed approaches.

\subsection{Fuzzy based Slip Suppression } \label{slip suppression}

\begin{figure}[!th]
 \removelatexerror
\begin{algorithm}[H]
\DontPrintSemicolon

  \KwInput{Incipient slip measures $(is_{e}$ and $is_{c})$ from feature based approach.}
   \KwOutput{Grip Force $g_{f}$}

   Fuzzification (\ref{fuzzify}) of inputs $is_{e}$ and $is_{c}$.\\
   Apply fuzzy operation (e.g.: AND fuzzy operator intersection (\ref{fuzzy_operation}) to evaluate the fuzzy rule (\ref{fuzzy_rule}) with multiple antecedents).\\
   Apply the Max-Min composition (\ref{reasoning_scheme}) reasoning scheme that involves clipping method and aggregation of the rule outputs.\\
   Defuzzify the aggregated output using COG technique (\ref{centroid_method}) and determine the grip force $\hat{g_{f}}$.\\
   \If {$\hat{g_{f}} > \hat{g_{f}}^{min} $}
{
 \If {$\hat{g_{f}} > \hat{g_{f}^{old}} $}
 {
${g_{f}} = \hat{g_{f}}(t)$.\\
 $\hat{g_{f}^{old}} = {g_{f}}$ 
}
   	}
   	\If {$\hat{g_{f}} > \hat{g_{f}}^{max} $}
 {
$g_{f} = \hat{g_{f}}^{max}$.\\
}
   return $g_{f}$.

\caption{Fuzzy based Slip Suppression}
\end{algorithm} \label{Fuzzy control algo}
\end{figure}

Fuzzy control strategies come from human expert experience and experiments rather than  from mathematical models. We utilize event-based slip data detected from the feature based approach to regulate the grip force to suppress slip. In general, a fuzzy logic controller consists of three segments namely fuzzifier, rule base and defuzzifier that implements the human heuristic knowledge.
We use mamdani type fuzzy controller to adjust the grip force using incipient slip feedback. The fuzzy based slip suppression method is summarized in Algorithm 3. We consider this problem as multi-input (two) and single output where a rule can be simply expressed as
\begin{equation}
\mathbf{IF}  \:\: is_{e} \:\: is \:\: A_{i} \:\:\: \mathbf{AND} \: \:\: is_{c} \:\: is \:\: B_{i} \:\:\: \mathbf{THEN} \:\: \:   g_{f} \: is \:\: C_{i},  \: i=1,...,n
 \label{fuzzy_rule}
\end{equation}
where $is_{e}$ and $is_{c}$ represents the accumulated number of edge and corner events from the initial time sample of a detected slip and $ g_{f}$ represents the grip force applied to control further slip. $A_{i}$ and $B_{i}$ are the input fuzzy sets, $C_{i}$ is the output fuzzy sets where i indicates the number of membership function.

In the fuzzification step, we first take the crisp feature inputs $is_{e}$ and $is_{c}$ and determine the degree to which these inputs belong to each of the appropriate fuzzy sets.
\begin{equation}
\mu_{A_{i}} : is_{e} \mapsto [0,1] \: ; \: \mu_{B_{i}} : is_{c} \mapsto [0,1] \:|
 \label{fuzzify}
\end{equation}
where the features is mapped to a value between 0 and 1.
In the rule based evaluation step, the fuzzified inputs are applied to the antecedents of the fuzzy rules. A given fuzzy rule has multiple antecedents where a fuzzy operation is used to evaluate the conjunction or disjunction of the rule antecedent. Since we follow \eqref{fuzzy_rule}, a fuzzy value $\delta_{i}$ output from an AND fuzzy operation intersection can be expressed as
\begin{equation}
\delta_{i} = \mu_{A_{i} \cap B_{i}} = min \{\mu_{A_{i}}(is_{e}), \mu_{B_{i}}(is_{c})\}
 \label{fuzzy_operation}
\end{equation}
The rule consequences are computed with respect to the inference mechanism. First, a clipping method is used to slice the consequent membership function at the level of the antecedent truth such that the rule consequent correlates with the true value of the rule antecedent. Then, the clipped membership functions of all rule consequent are combined in to a single fuzzyset. This process of unification of the output of all rules that is easier to defuzzify are otherwise called aggregation. Finally, the Max-Min composition generates an aggregated output surface:
\begin{equation}
\mu^{c}(g_{f}) = max_{i} (min (\delta_{i}, \mu_{C_{i}}(g_{f}))
 \label{reasoning_scheme}
\end{equation}
In the defuzzification step, the aggregated output fuzzy set goes through centroid method and outputs a single grip force value. The centroid method  determines a point representing the centre of gravity (COG) such that when a vertical line drawn at the point could split the aggregate set in to two equal masses. The COG can be mathematically expressed as 

\begin{equation}
g_{f}  = f(is_{e}, is_{c}) = \frac{\int_{a}^{b} g_{f}\mu^{c} (g_{f}) dg_{f} }{\int_{a}^{b} \mu^{c} (g_{f}) dg_{f}}
 \label{centroid_method}
\end{equation}
The controller generates an output only when there is a increase of grip force and that is within the specified limits. Robotic fingers applies such grip force increments and suppress slip.

\section{Experimental Settings }
In this section we describe the experimental setup and protocol used to conduct slip detection and suppression experiments. 
\subsection{Experimental Setup }
%


\begin{figure}[h!]
 \centering
            \subfloat{\label{fig:sf1}
      \includegraphics[width=0.45\textwidth, height=0.25\textwidth]{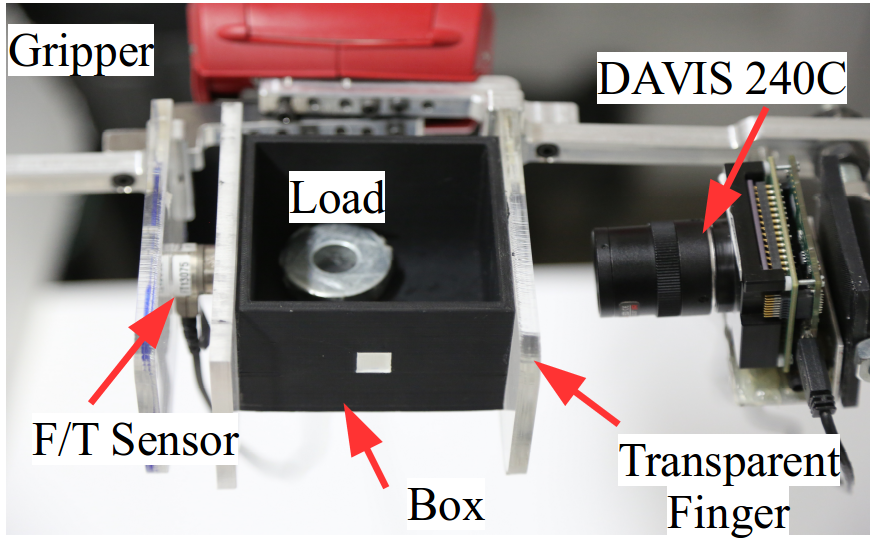}}  \hfill \\
\caption{Experimental Setup: Baxter robot holding a lifted box object using the event-camera based finger prototype.}
\label{setup_process}
\end{figure}


The experimental setup consists of Baxter robot, Electric parallel gripper, F/T sensor (ATI Nano17), Dynamic and active pixel vision sensor (DAVIS240C), event-camera based finger prototype and box object shown in Fig. \ref{setup_process}.

Baxter is a dual-arm compliant robot, each arm having seven joints and a electric parallel gripper designed mainly to handle tasks in production line and human centered environments. The parallel electric gripper provide one degree of freedom and has different opening positions starting from 5 \% to 95 \% which corresponds to a distance between the gripper sides, 11.7 cm to 14.7 cm respectively. Moreover, the gripper allows both position and force control. Furthermore, the clips that fit over the gripper base to handle different size (0-15 cm) of objects facilitates the attachment of custom made fingers.

In our experiments, we replaced the existing fingers of the Baxter gripper with our newly designed finger for grasping objects. The finger vision prototype shown in Fig.\ref{setup_process} (a-(2)) has two metallic frames with adjustable camera slots and fixed transparent acrylic plates where a F/T sensor and an event-based camera is integrated. In particular, the F/T sensor is placed in one side of the gripper in a sandwich arrangement between two acrylic plates to monitor the grip force and force changes due to the slip of the grasped object. We used ATI Nano17 F/T sensor which is one of the smallest, light weight and high resolution 6-axis transducer commercially available that can resolve down to 0.318 gram-force. This transducer is connected to a net F/T system which measures six components of force and toque and communicates with a host computer through a high speed Ethernet interface. Moreover, the F/T sensor mainly serves the purpose to validate the slip detected from the  DAVIS 240C camera.

On the other side of the gripper, DAVIS 240C camera with a c-mount lens is mounted at the backside of the acrylic plate to detect the object slip and to provide feedback to the gripper for slip control. The DAVIS 240C combines both frame (active pixel sensor APS) and event (DVS) based sensor with a pixel level resolution of 240 $\times$ 180. It has a minimum latency of 12 $m s$, bandwidth of 12 MEvent/second and a dynamic range of 120 dB which is connected to a host computer through a USB 2.0 cable.

\begin{figure*}[th!]
  \centering
      \subfloat[]{\label{fig:sf1}
      \includegraphics[width=0.32\textwidth, height=0.35\textwidth]{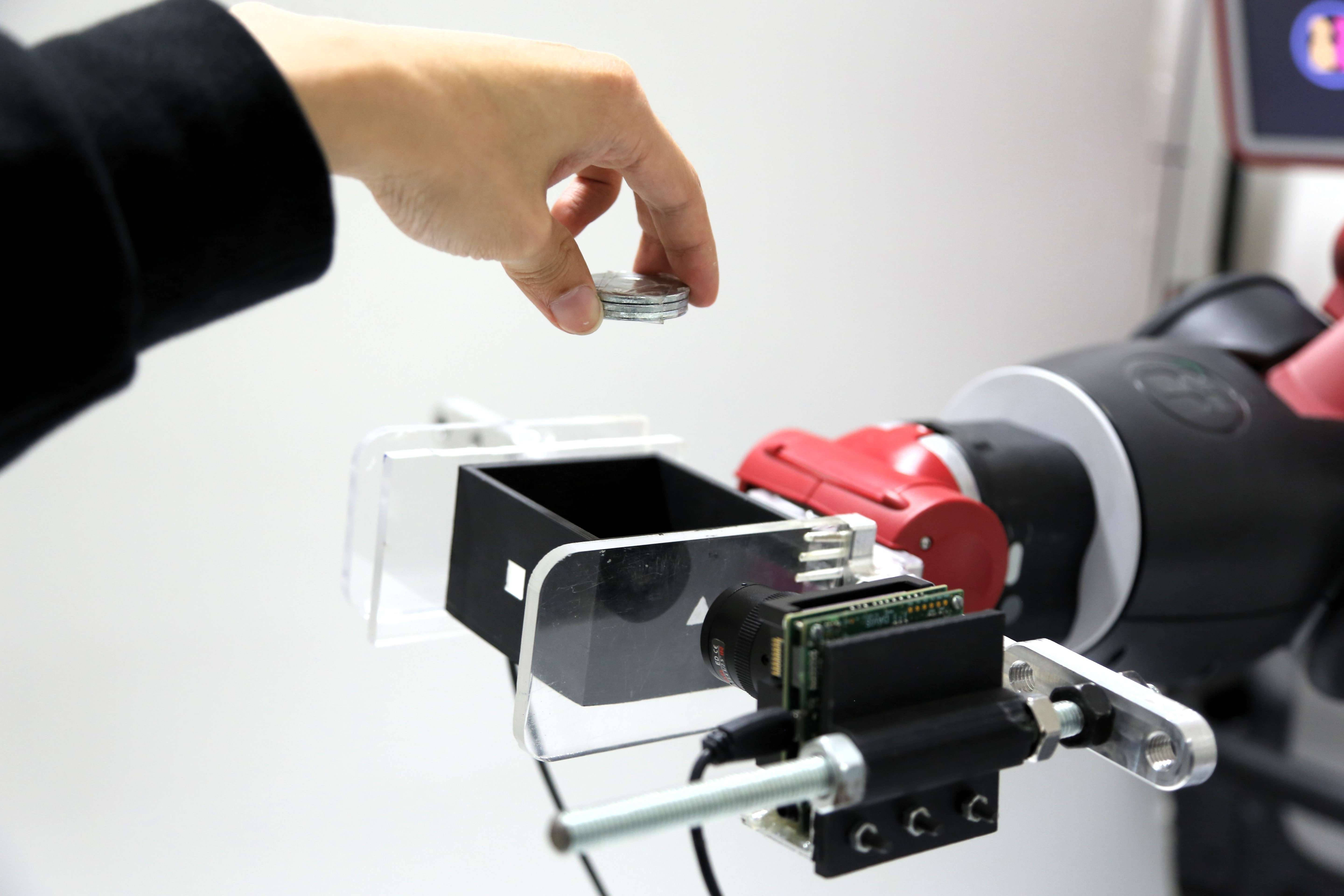}}      \hfill
      \subfloat{\label{fig:sf1}
      \includegraphics[width=0.63\textwidth, height=0.35\textwidth]{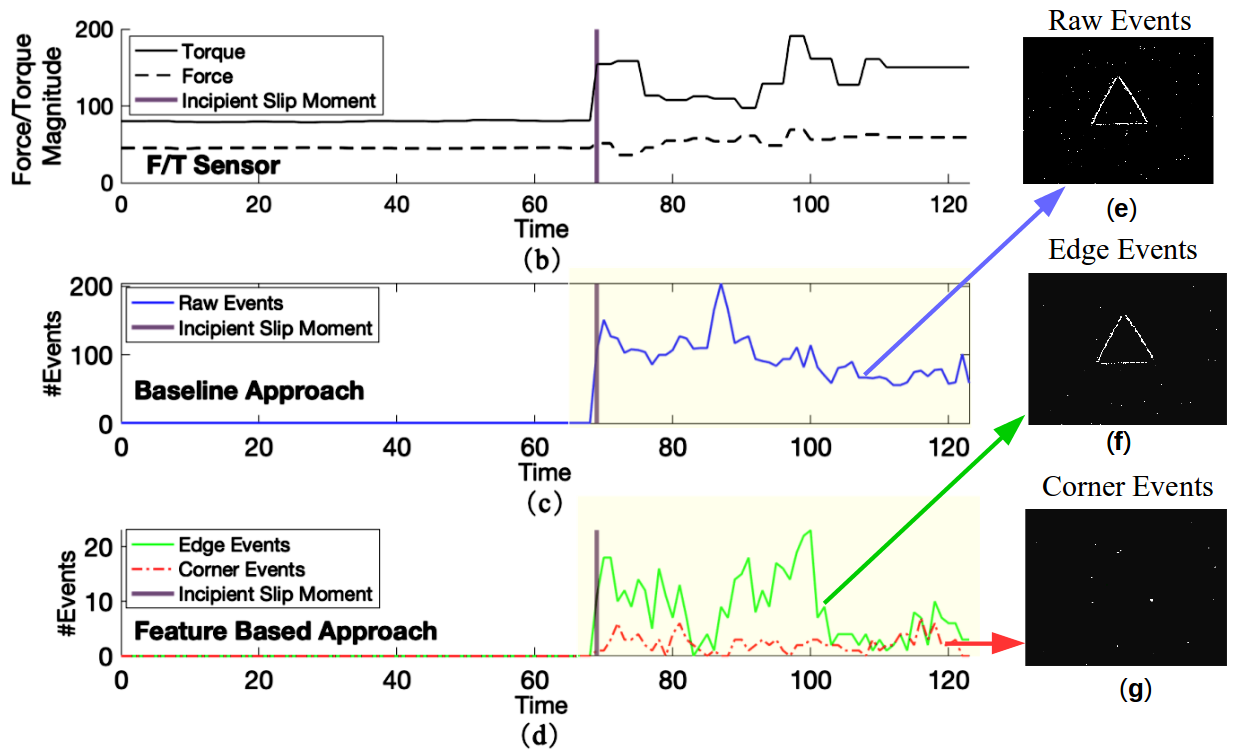}} 

      \hfill
\caption{Baseline vs Feature based approach: (a) Human hand posed to drop weight and induce slip on the grasped object. (b) F/T sensor records the changes caused by the slip. (c) Baseline approach detects object slip by directly analysing the raw events. (d) Feature based approach analyses the features such as edges and corners to determine slip. (e-g) rendered event-based frame corresponding to the slip signals from baseline and feature based approaches. Both approaches operates at 500 $\mu s$ and uses online sampled noise as thresholds. }
\label{Baseline_feature_Approach}
\end{figure*}
Middleware is crucial for multi-robots/hardware operation and communication. Robot Operating System (ROS) is a popular software framework and middleware for writing robotic applications. ROS is preferred due to their inter-platform operability, modularity, the core value of code reuse, active development of hardware drivers and application software by the research community. We set up the robot and independent hardware (Force/Torque Sensor, Dynamic Vision Sensors) in ROS for the development of slip detection and control algorithms. In particular, the Baxter PC and host computer operates on Ubuntu 16.04 with ROS kinetic version. This setting allows clean and reliable operations as well as extraction of repeatable data. Moreover, the rosbag feature in ROS which allows recording the complete experimental data, enables us to test slip detection and control algorithms off-line and compare their performance and validate their effectiveness with ground truth information from other sensors.

A square box carved with primitive marker on each side is used in the experiments. Especially, the object is placed in between the fingers in such a way that the marker faces the camera side of the gripper. To induce object slips under grasped condition, a light (80 grams) and heavy (200 grams) weight block is used.

\subsection{Experimental protocol } \label{exp_protocol}


In each experiment, we followed a three stages of process to conduct robot grasping and manipulation task while enabling slip detection and suppression. In the following the stages are explained in detail.

Sampling Stage: The DAVIS240C camera is sensitive to brightness change from the surrounding environment. This generates sparse data (noise events) without any occurrence of relative motion between object and camera. In this stage, we sample the noise by accumulating the number of events for a rolling temporal window (eg: 0.5 $ms$ or 10 $ms$) chosen for the slip detection algorithm. Mechanical vibrations and illumination uncertainties in static condition are also captured in the sampling process. The maximum number of events obtained from the uniform time samples is used as a threshold to distinguish noise and events from moving object. 

Grasping Stage: A grasp pose is determined for a known object in such a way the object is placed between the fingers and the DVS camera is able to fully observe the marker events within the finger boundaries. Then, the robot manipulator does motion planning and reaches the pre-grasp pose. Finally, the gripper executes the grasp with minimal grip force or does caging of the object.


Slip Monitor Stage: Slip incidents that occurs while performing a given robotic object manipulation task that includes grasping, lifting, loading, lowering and dropping are detected and suppressed at this stage. For example, under grasped condition, weight is added on the object to induce slip. The object slides when the added weight exerts a force greater than the friction force between the gripper and the object. The triggered events are monitored and slip incidents are detected based on the proposed approaches. Earlier, we presented two approaches, one baseline and another a feature based in Section. \ref{slip_detection} to detect slip using event-based finger vision system. The approaches detect slip under uncertainty and measures from the F/T sensor are used to validate the actual slips shown in Fig. \ref{Baseline_feature_Approach} (b). Moreover, the fuzzy based grip force control presented in Section. \ref{slip suppression} is applied to suppress the object slips based on the feature-based slip detection. 
%


The proposed approaches takes the stream of events from the DAVIS 240 C camera and directly process raw events at a sampling rate of $500 \mu s$ to detect incipient and gross slip. The sampling rate is chosen such that the magnitude of events are significant and noise has less influence. In all of our experiments, the labeled stream of events goes through all three stages and processed accordingly. In Fig. \ref{Baseline_feature_Approach} (c), the baseline approach using the direct raw events goes through all three stage of process, first the sampling stage determines a noise threshold, second the object is grasped and a weight is added. The incipient slip (spike), gross slip (signal appear in the yellow region) and event-based frame are illustrated in (c) and (e). The feature-based approach takes the stream of events from the DAVIS 240 C camera and labels them as edge and corner events.  In Fig. \ref{Baseline_feature_Approach} (d), the detected incipient slip and gross slip due to an added weight and further object motions are illustrated with corresponding features. Moreover, a slice of an event-based frame is shown for each labeled events in (f) and (e).

\section{Slip Detection Experiments and Results}

\subsection{Event-based Slip Detection Accuracy and F/T validation}

Several experiments were conducted to determine the efficacy of the baseline and feature-based slip detector for use in the slip suppression strategy. Especially, the slip detection accuracy of both approaches are tested under object grasped conditions by adding weight. For each experiment, we observe whether a slip is detected early or after or at the moment of the actual induced slip and these criteria are classified accordingly:

1) A true positive and true negative is recorded when the slip is detected at the exact moment of actual induced slip. 

2) A false positive and false negative recorded when a slip is detected earlier before the actual slip or remains undetected after the actual slip.
The actual slips are validated by the changes in F/T sensor which is an integrated part of event-based finger vision system.

Pre-experiment procedure: The left-arm end effector of the Baxter robot reaches the pre-grasp pose determined from grasp planning and executes the grasp. The gripper was commanded to hold the box object with a static force of 5N at each contact point. The holding force is determined from earlier experimental lift trials. Then, the manipulator lifts the grasped object straight up (50 cm) at a constant speed and stationed. 

Load test procedure: After this sequence, a human user position his hand to drop the load (200 gm) from a height ranging between 4cm to 8cm above the grasped object shown in Fig. \ref{Baseline_feature_Approach} (a). The slip accuracy test is conducted by dropping weight on the grasped object. We recorded the slip signals from the detector as well as the F/T measures for actual slip validation and compared to the above classifications. 

\begin{table}\caption{Confusion Matrix of Event-Based Slip Detectors}\centering\ra{1.3}\begin{tabular}{@{}rrrrcrrrcrrr@{}}\toprule& \multicolumn{2}{c}{Baseline Approach} & \phantom{abc}& \multicolumn{2}{c}{Feature-based Approach} \\\cmidrule{2-3} \cmidrule{5-6} & Slip & No-Slip  && Slip & No-Slip \\\midrule Slip & $80 \% $ & $20 \% $ && $98 \% $ & $2 \% $ \\No-Slip &  $20 \% $& $80 \% $&& $2 \% $& $98 \% $ \\\bottomrule\end{tabular}\label{confusion_matrix}\end{table} 
Results: Fig. \ref{Baseline_feature_Approach} (b-d) illustrates the incipient slip and gross slip signals detected by the proposed approaches under controlled environment, F/T measures to validate them and images depicting the accumulated slip events corresponding to the approaches. The baseline approach directly use raw events in slip detection and accounts the noise in event stream, (e) depicts the  noise and actual slip events in an event-based frame. The feature based approach employ event-based feature detector (e-Harris) to detect corners and edges from the raw events and use them in slip detection. Images, (f) and (g) illustrates the accumulated corner and edge events belonging to the slip signal emphasizing informative events and the presence of less noise. The experiments were repeated 50 times for both baseline and feature based slip detector and for each repetition the accuracy is evaluated with respect to the classification.

The experimental results were compiled in the form of confusion matrix shown in Table. \ref{confusion_matrix}. Only one repetition for feature based approach detected false slip whereas baseline approach detected false slips, ten out of fifty experiments. The accuracy of the feature based slip detector is very high compared to the baseline detector. Even though both approaches perform better in controlled and ideal setting, small vibrations and varying illumination are common in compliant robot manipulation. In the next experiment, we examine the robustness of slip detectors under such uncertainties comprehensively.

\subsection{Testing Robustness of the Slip Detection Approaches under Uncertainties} \label{noise_level_analysis}

\begin{figure}[ht]
  \centering
      \subfloat{\label{fig:sf1}
      \includegraphics[clip,trim = 1.5cm 1cm 0cm 0.5cm, keepaspectratio,width=0.46\textwidth, height=2\textwidth]{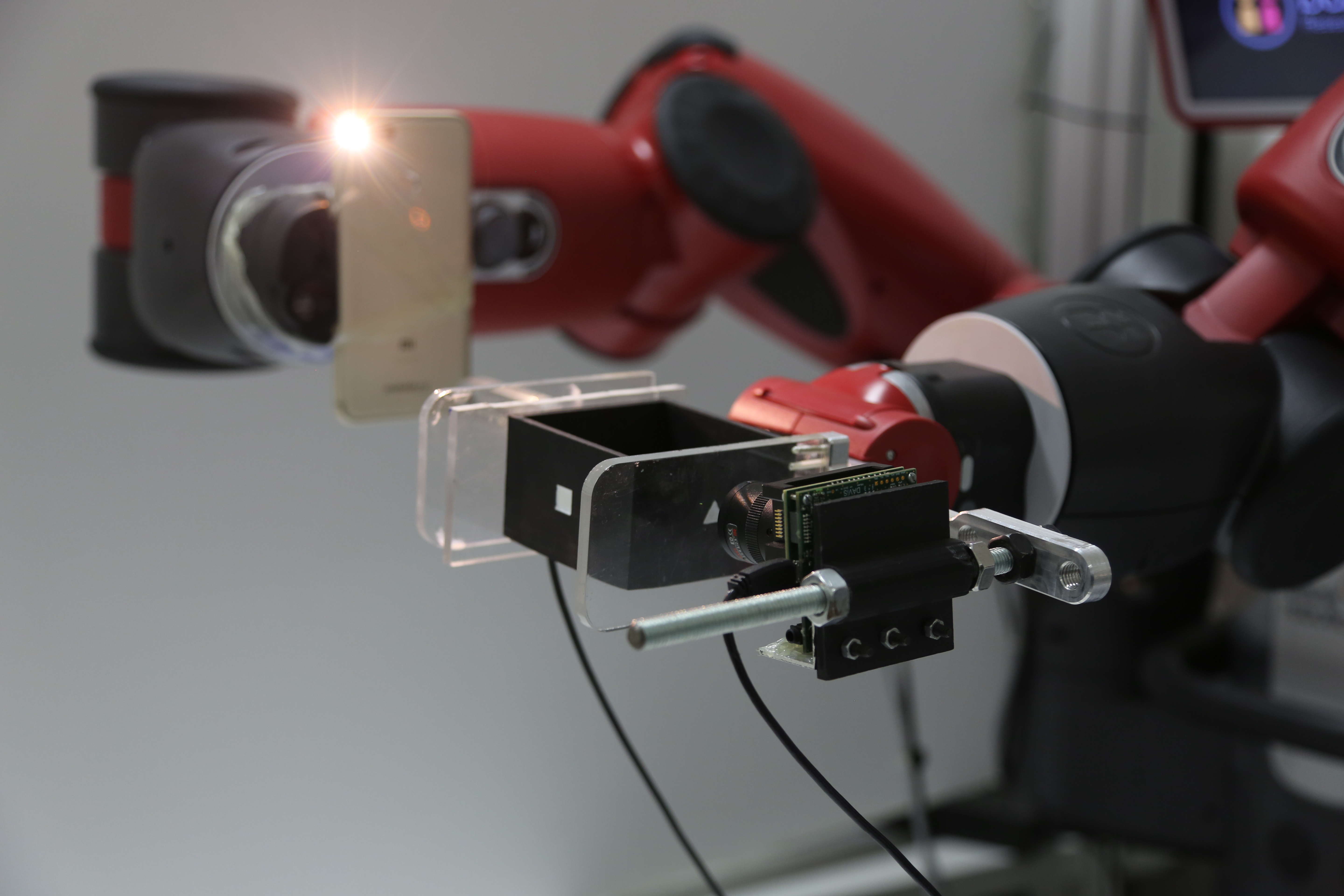}} 
      \hfill 
      
\caption{Baxter right arm end effector with the light source moves to vary illumination in the experimental setting}
\label{baxter_light}
\end{figure}

\begin{figure}[th!]
  \centering
      \subfloat[]{\label{fig:sf1}
      \includegraphics[clip,trim = 1.5cm 1cm 0cm 0.5cm, keepaspectratio,width=0.5\textwidth, height=2\textwidth]{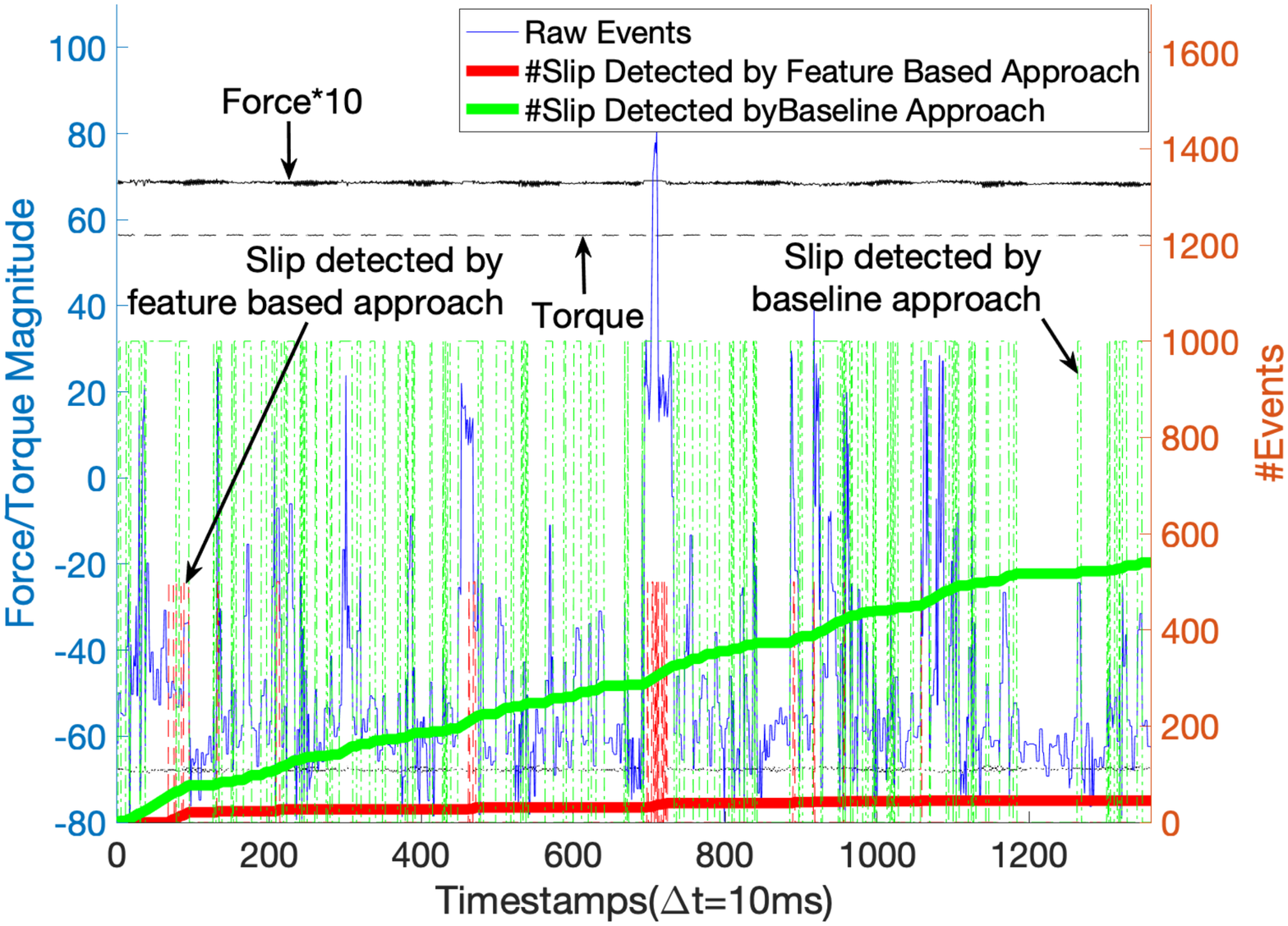}} 
      \hfill \vspace{-1em}
      \subfloat[]{\label{fig:sf1}
      \includegraphics[clip,trim = 1.5cm 1.5cm 0cm 0.5cm, keepaspectratio,width=0.5\textwidth, height=2\textwidth]{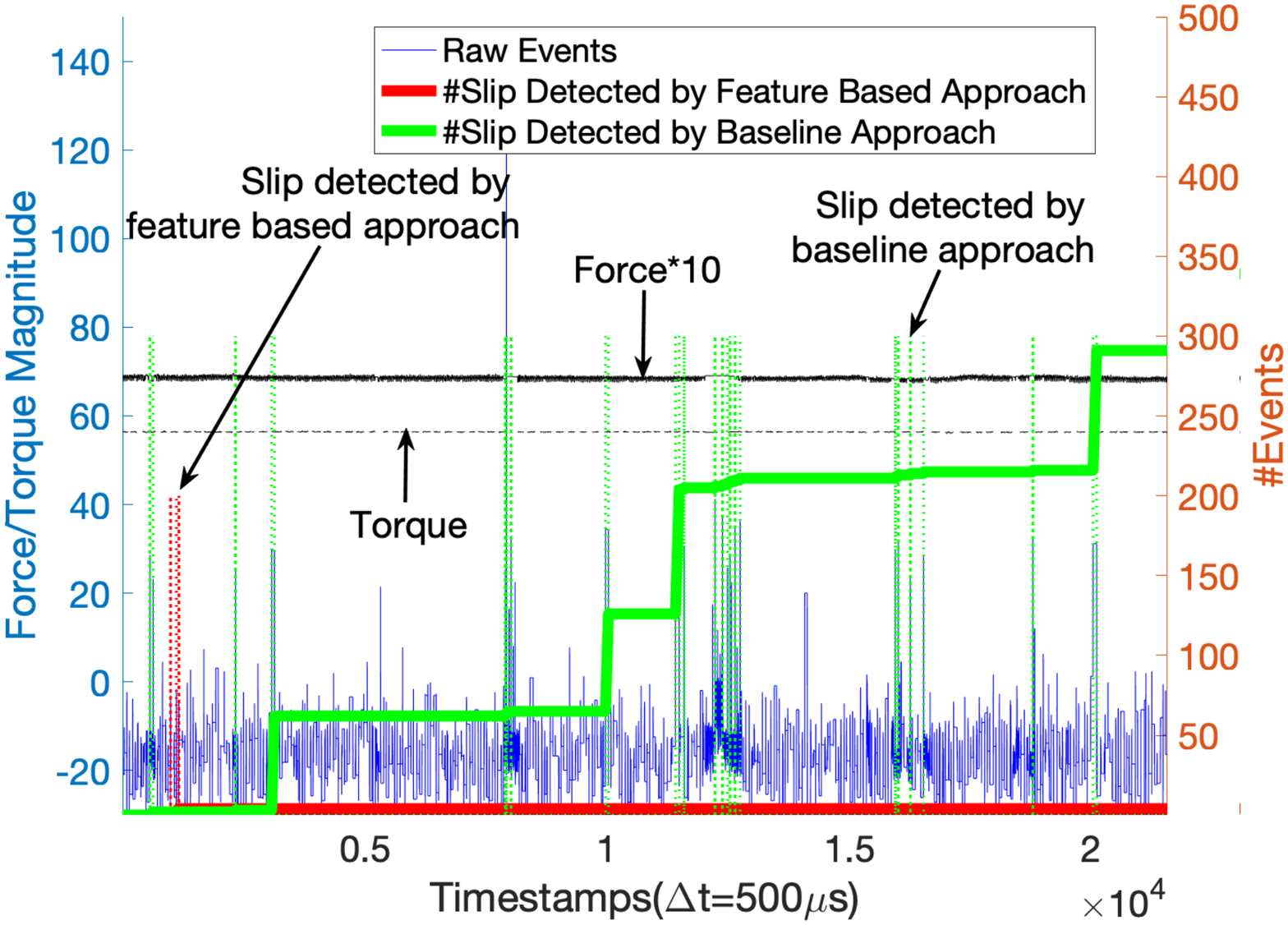}} 
\caption{Approaches performance at 10 ms (a) and 500 $\mu s$ (b) time sampling: False slips detected by Baseline and Feature based approach for the noise events generated by varying illumination and small vibrations.}
\label{rbf}
\end{figure}
\subsubsection{Baseline and Feature-based Approach performance under uncertainty}

To examine the robust slip detection ability of the approaches, we conducted experiments with two different sampling rates (0.5 ms and 10 ms) for detecting false slips under varying lighting conditions and small vibrations from the robot compliance. 

Illumination test procedure: In these experiments, we mounted a white LED light source in the right-arm end effector of Baxter robot to artificially induce illumination changes to experimental environment shown in Fig \ref{baxter_light}. Following the pre-experiment procedure, the left-arm with the grasped object was positioned 25 cm away from right-arm end effector. The right arm was moved in the x-z plane in the sequence of up, down, right and left moving back and forth from center. Raw events from the event camera and false slip signals from the detector got recorded for a period of time. This

 In Fig. \ref{rbf}, the slip detection performance of the approaches from an experiment is illustrated for 0.5 ms and 10ms temporal sampling. The baseline approach detected a huge number of false slips caused by the lighting noise and small vibrations whereas the feature based approach demonstrated its robustness by detecting only few false slips. The baseline approach improved its performance by two-fold for a smaller sampling rate. The combination of corner and edge features used in feature based approach tackled noise events equally well in both sampling rates. We conducted sixteen experiments to analyse the robustness of the approaches under different the two different sampling rates. The robust ability of the baseline and feature-based slip detector improved by 50 \%  and 20 \% with smaller (0.5 ms) sampling rate. Therefore, 0.5 ms temporal sampling is used in further analysis.
\begin{figure}[th]
  \centering
      \subfloat[Set point 1 and 2 captures the range of noise events]{\label{fig:sf1}
      \includegraphics[clip,trim = 2.8cm 1.5cm 0cm 0.5cm, keepaspectratio,width=0.54\textwidth, height=1\textwidth]{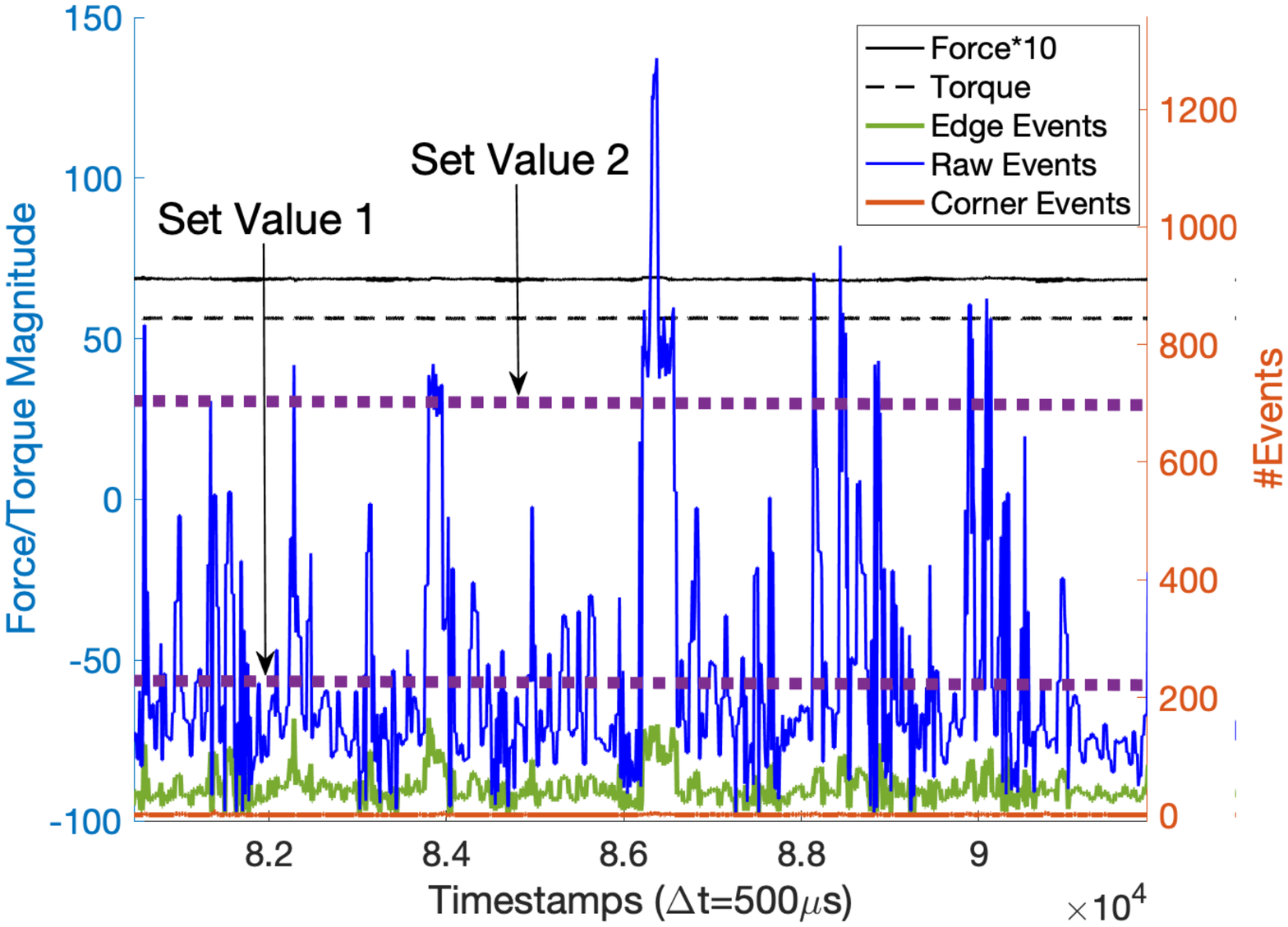}} 
      \hfill
      \subfloat[Probability of individual features misdetecting slips under noise condition]{\label{fig:sf1}
      \includegraphics[clip,trim = 0cm 0cm 0cm 0cm, keepaspectratio,width=0.40\textwidth, height=1\textwidth]{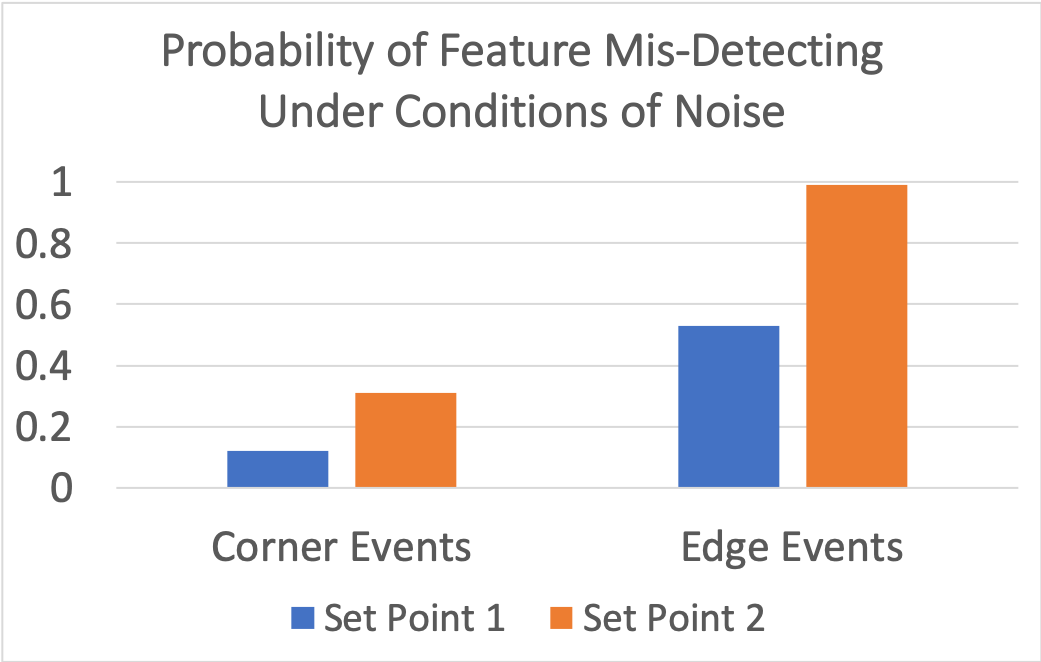}} 
\caption{Noise levels and its effect in feature based approach}
\label{Noise level comparison}
\end{figure}
\begin{figure}[th]
  \centering
      \subfloat{\label{fig:sf1}
      \includegraphics[clip,trim = 2.5cm 1.5cm 0cm 0.5cm, keepaspectratio,width=0.52\textwidth, height=1.4\textwidth]{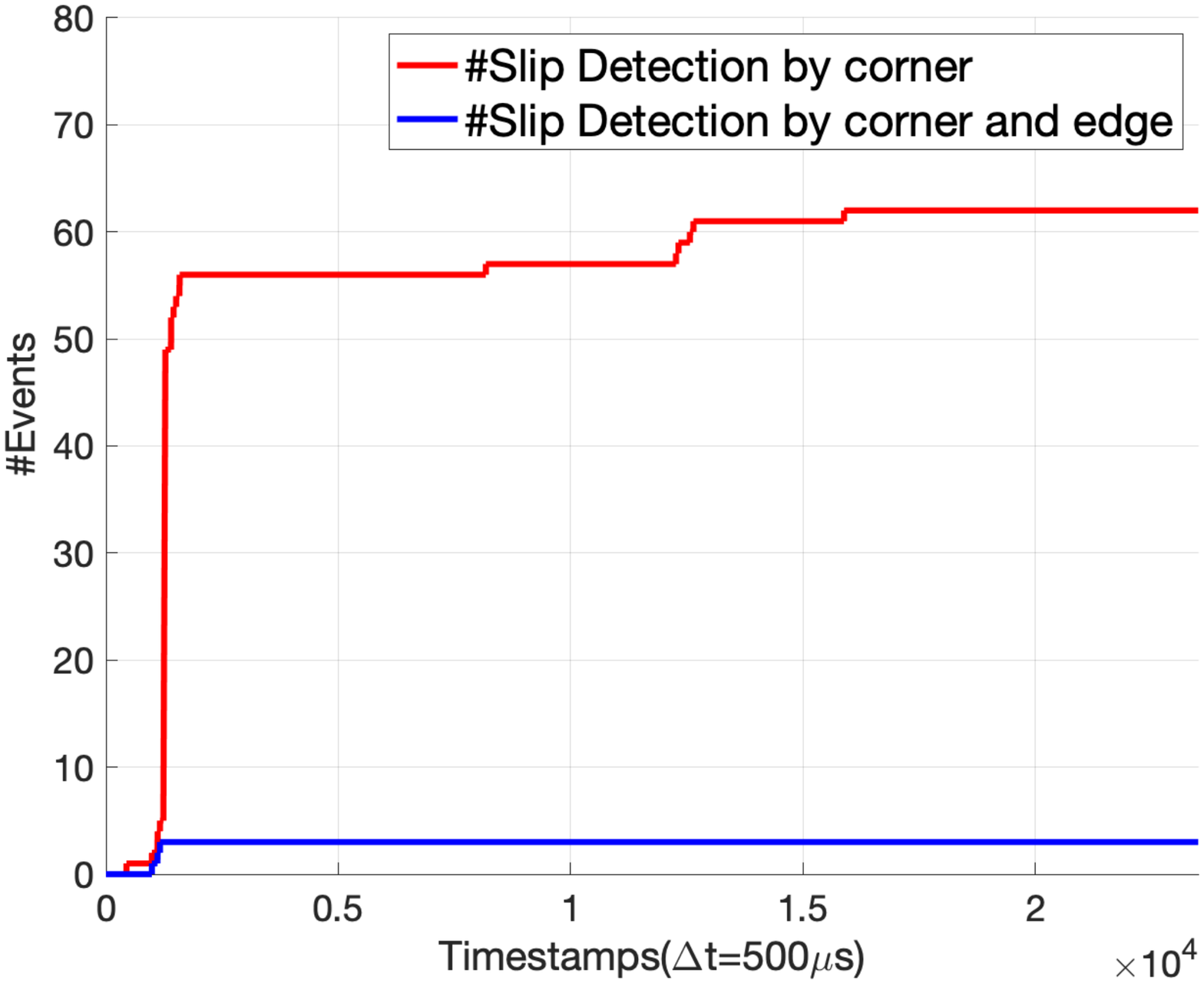}}  
      \caption{Feature-based Approach: Single and multiple features used to detect slip at 500 $\mu s$ time sampling.}
\label{bvsf}\vspace{-2em}
\end{figure}
We also studied how the level of noise induced by the varying illumination in the experimental setup affect the feature based slip detector. The noise measured from the sampling stage under controlled lighting conditions is taken as the base threshold. The further two set points are three times and six times of the base threshold. In Fig. \ref{Noise level comparison} (a), the labeled stream of events such as the raw, corner and edges and three set points indicating the noise level are illustrated from an experiment. In particular, the varying raw events represents the illumination uncertainty and featured events reflecting their impacts. The mean value of feature events corresponding to the noise level set points indicates that the corner feature is more robust than the edge feature for different noise levels. Experimental results from sixteen experiments shown in \ref{Noise level comparison} (b) indicates the corner feature has 10 \% and 50 \% chance of detecting false slip for set point 1, whereas the for increased noise (set point 2) there is 50  \% and 100 \% chance for false slip detection.

Addition to the earlier study, we examine the ability of the feature based approach using single and multiple features for detecting slips under noise and vibration uncertainty. The multiple-feature combination detected less false slips compared to single feature based slip detection shown in Fig. \ref{bvsf}. Thus, using multiple features increases the robust slip detection ability of the feature based approach.


\subsubsection{Comparison of Event-based Corner Detectors under Uncertainty}

The state of the art event-based corner detectors were used in the feature based slip detection approach and their robust performance in tackling illumination and vibration uncertainty is compared.

\begin{figure}[!th]
  \centering
      \subfloat[]{\label{fig:sf1}
      \includegraphics[clip,trim = 2.5cm 1.5cm 0cm 0.5cm, keepaspectratio,width=0.52\textwidth, height=2\textwidth]{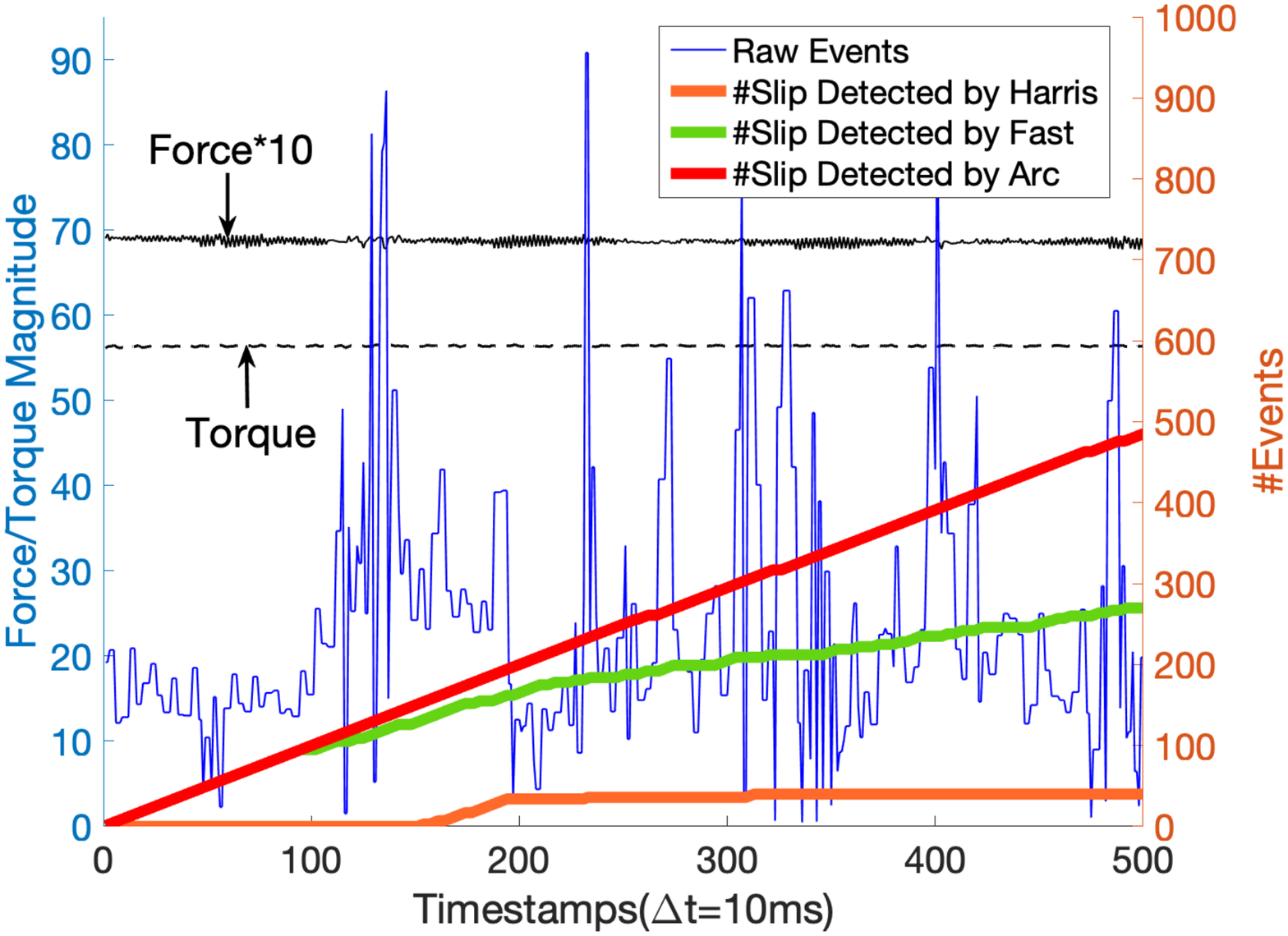}} 
      \hfill \vspace{-2em} \\
      \subfloat[]{\label{fig:sf1}
      \includegraphics[clip,trim = 2.8cm 1.5cm 0cm 0.5cm, keepaspectratio,width=0.55\textwidth, height=2\textwidth]{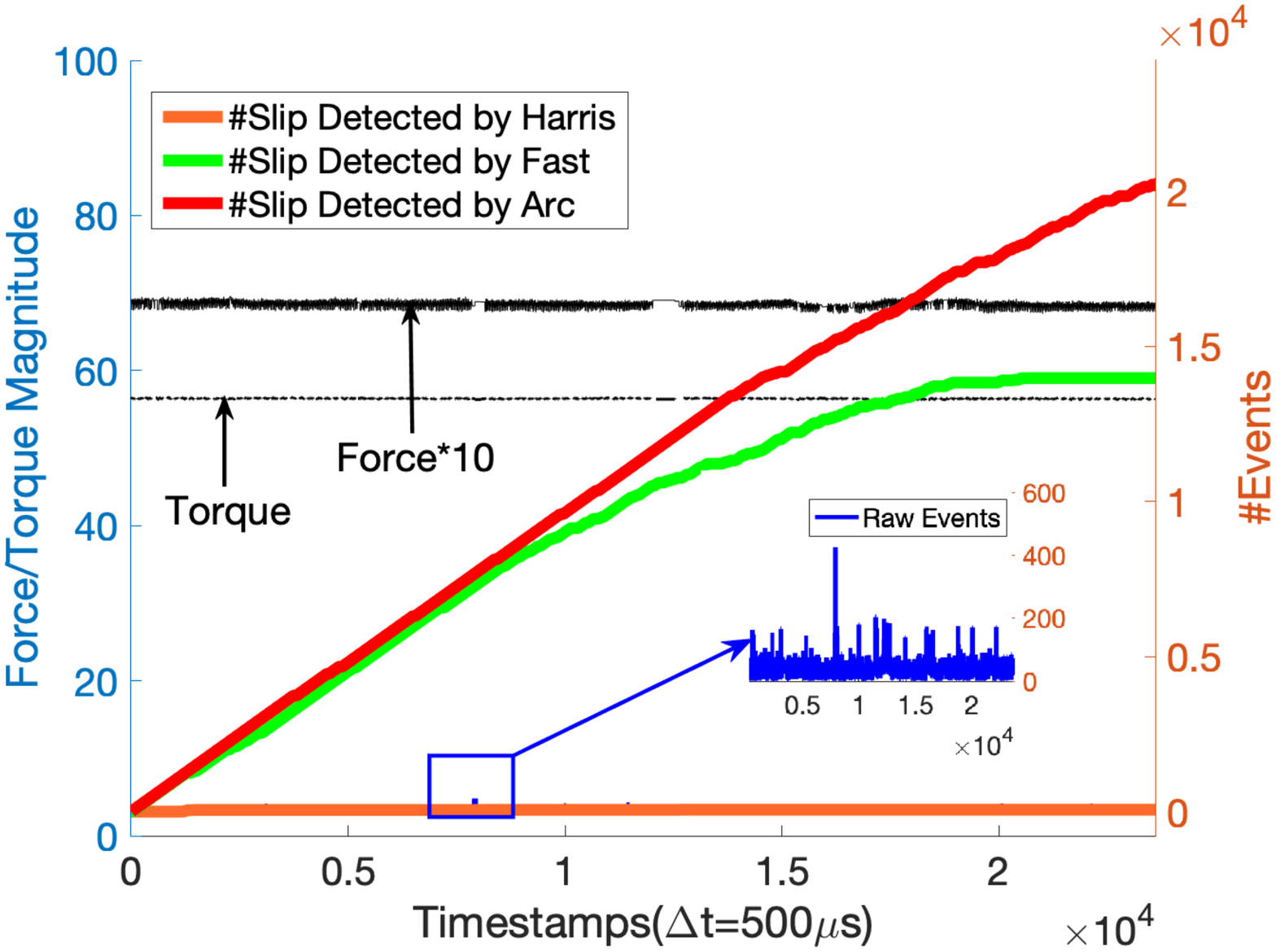}} 
\caption{Comparison of event-based corner detectors via feature based approach for robust slip detection at 10 ms (a) and 500 $\mu s$ (b) time sampling: False slips detected by feature based approach using e-Harris, FAST and ARC corner detectors under varying illumination and small vibrations.}
\label{Corner detector comparison}\vspace{-1em}
\end{figure}

We conducted three experiments to test the effectiveness of the three corner detectors which is e-Harris, e-Fast and ARC* in tackling the noise events in slip detection at two sampling rates. For a period of time shown in Fig. \ref{Corner detector comparison} (a) and (b), the e-Harris detected few false slips and showed robust performance in both sampling rates. Surprisingly, the e-Harris and ARC* corner detector performed poorly at lower sampling rate which demonstrates their poor ability to withstand noises caused by varying illumination and small vibration. For both sampling rates, the more efficient corner detector showed worse performance in terms of accuracy. In particular, ARC* performed worse than the efficient e-fast consistently. The e-Harris shows superior accuracy performance over the other two corner detectors. Therefore, we adopt e-Harris method and utilize multiple features for slip detection and suppression.

\subsubsection{Detection of Actual Slips under uncertainty }
%

\begin{figure}[h!]
  \centering
      \subfloat[F/T measures and slips detected by the proposed approaches]{\label{fig:sf1}
      \includegraphics[clip,trim = 1.5cm 1cm 0cm 0.5cm, keepaspectratio,width=0.5\textwidth, height=1.5\textwidth]{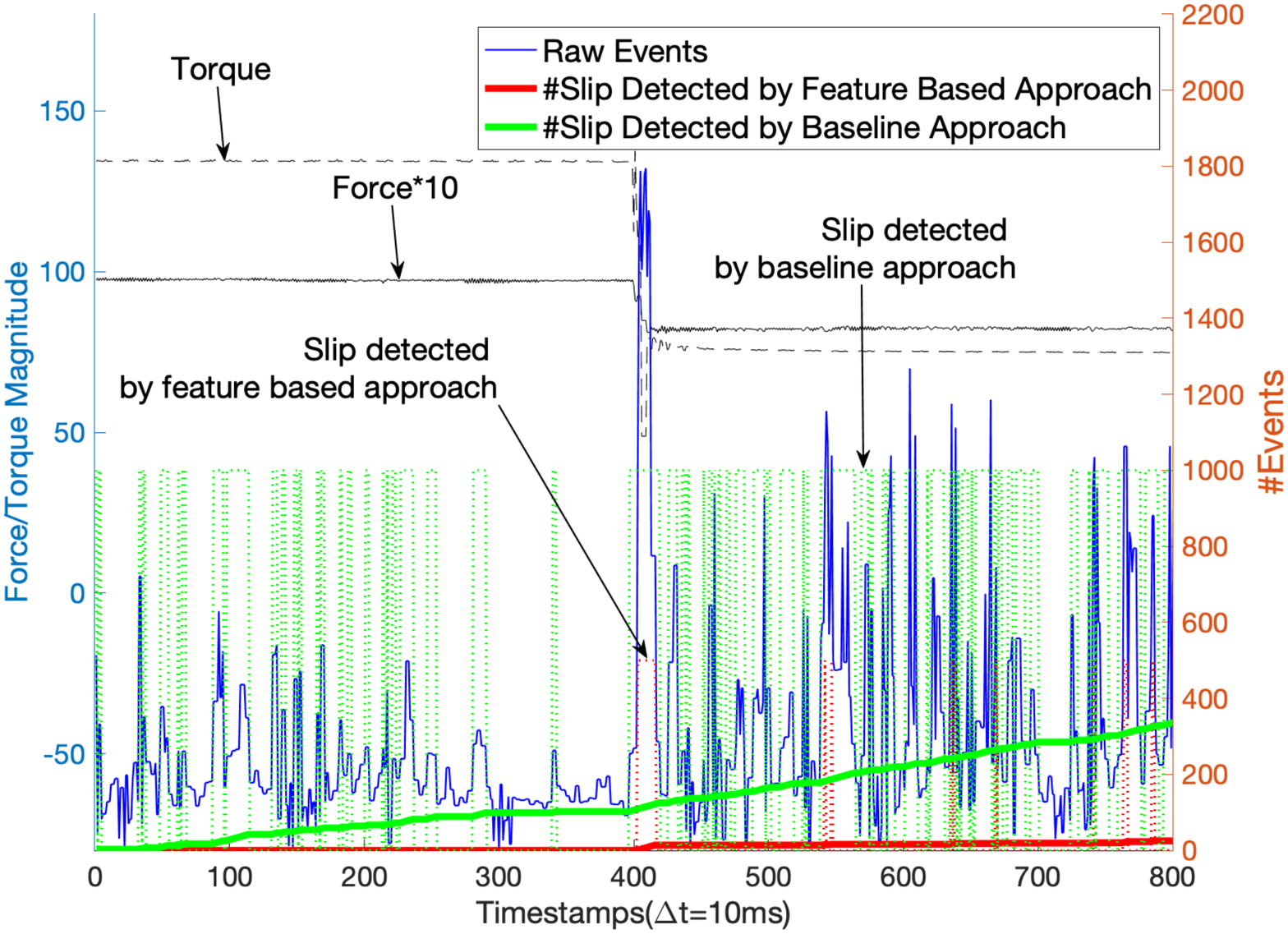}} 
\caption{Detection of actual and false slips by the proposed approaches under illumination and vibration uncertainty.}
\label{load_noise_test}
\end{figure}

We study the viability of the approaches in detecting actual slips under varying illumination and small vibrations. We performed twenty five experiments where each experiment follows the illumination-test and load-test procedure for testing the effectiveness of the proposed slip detection approaches. In simpler terms, we induce slip by adding weight on the grasped object under vibration and illumination uncertainty. In these experiments, we decreased the sensitivity of slip detectors by 10 \% which improved the rejection of false slips and detection of actual slip.

For a load added on the grasped object, Fig. \ref{load_noise_test} depicts the actual induced slip and false slips detected by the baseline and feature based approach, F/T measures validating the actual slip. The success rate from the approaches are compiled in a confusion matrix shown in Table. \ref{confusion_matrix_noise}. Both the approaches are able to detect object slips. However, the baseline approach fails to be robust against noises and unable to distinguish false and actual slips, succeeds only four out of twenty five experiments. Only two repetition detected false slips for the feature based approach. In the following session, we utilize the feature-based approach for detecting slip signals and feedback them to adjust the grip force such that the slip is suppressed.


\begin{table}\caption{Confusion Matrix of Event-Based Slip Detectors under Vibration and Illumination Uncertainty}\centering\ra{1.3}\begin{tabular}{@{}rrrrcrrrcrrr@{}}\toprule& \multicolumn{2}{c}{Baseline Approach} & \phantom{abc}& \multicolumn{2}{c}{Feature-based Approach} \\\cmidrule{2-3} \cmidrule{5-6} & Slip & No-Slip  && Slip & No-Slip \\\midrule Slip & $16 \% $ & $84 \% $ && $92 \% $ & $8 \% $ \\No-Slip &  $84 \% $& $16 \% $&& $8 \% $& $92 \% $ \\\bottomrule\end{tabular}\label{confusion_matrix_noise}\end{table}

\begin{figure}[h!]
  \centering
      \subfloat[Edge events based slip patterns]{\label{fig:sf1}
      \includegraphics[clip,trim = 2.5cm 2.5cm 0cm 1.5cm, keepaspectratio,width=0.55\textwidth, height=1.5\textwidth]{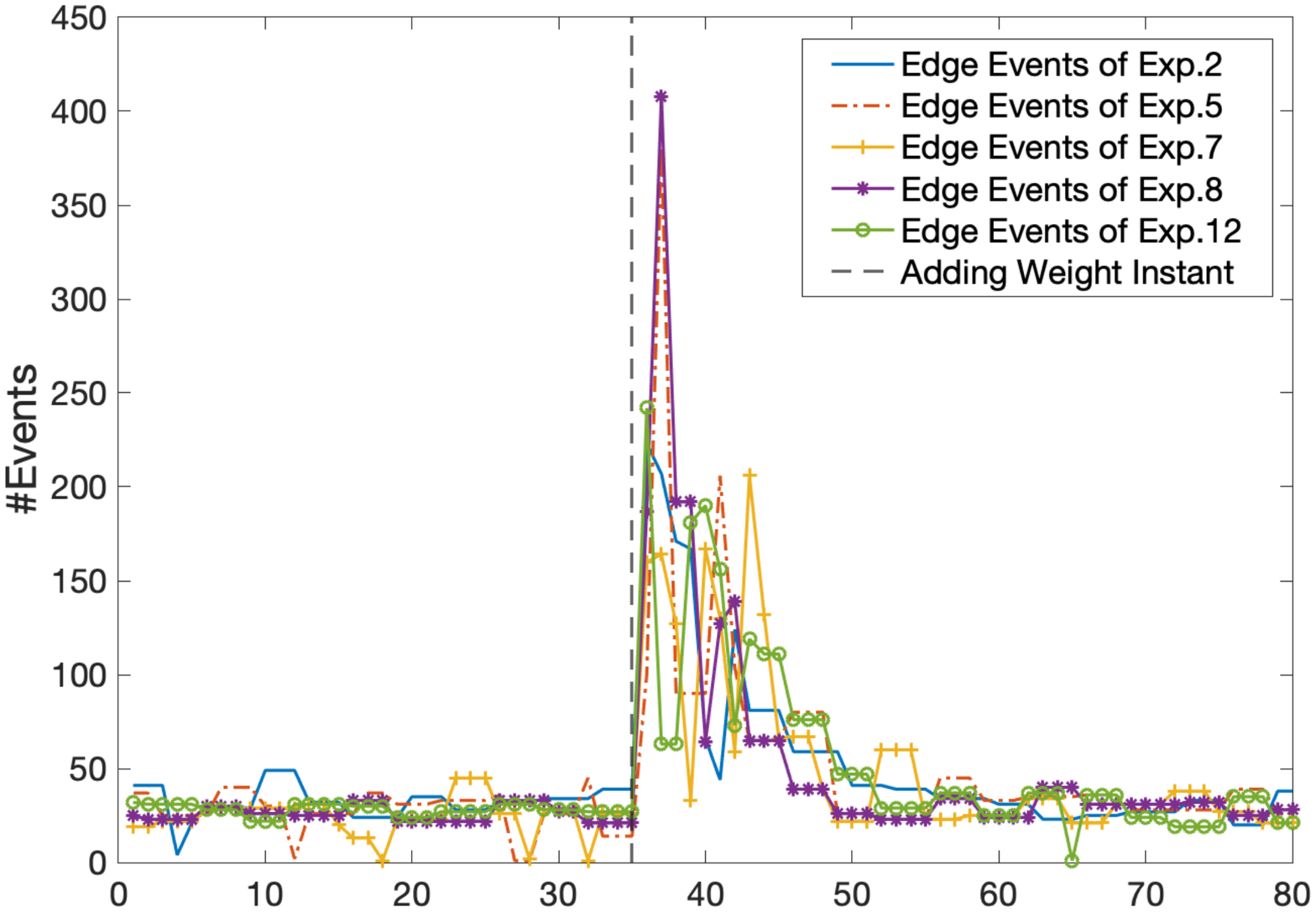}} 
      \hfill
      \subfloat[Corner events based slip patterns]{\label{fig:sf1}
      \includegraphics[clip,trim =  2.5cm 2.5cm 0cm 1.5cm, keepaspectratio,width=0.55\textwidth, height=1.5\textwidth]{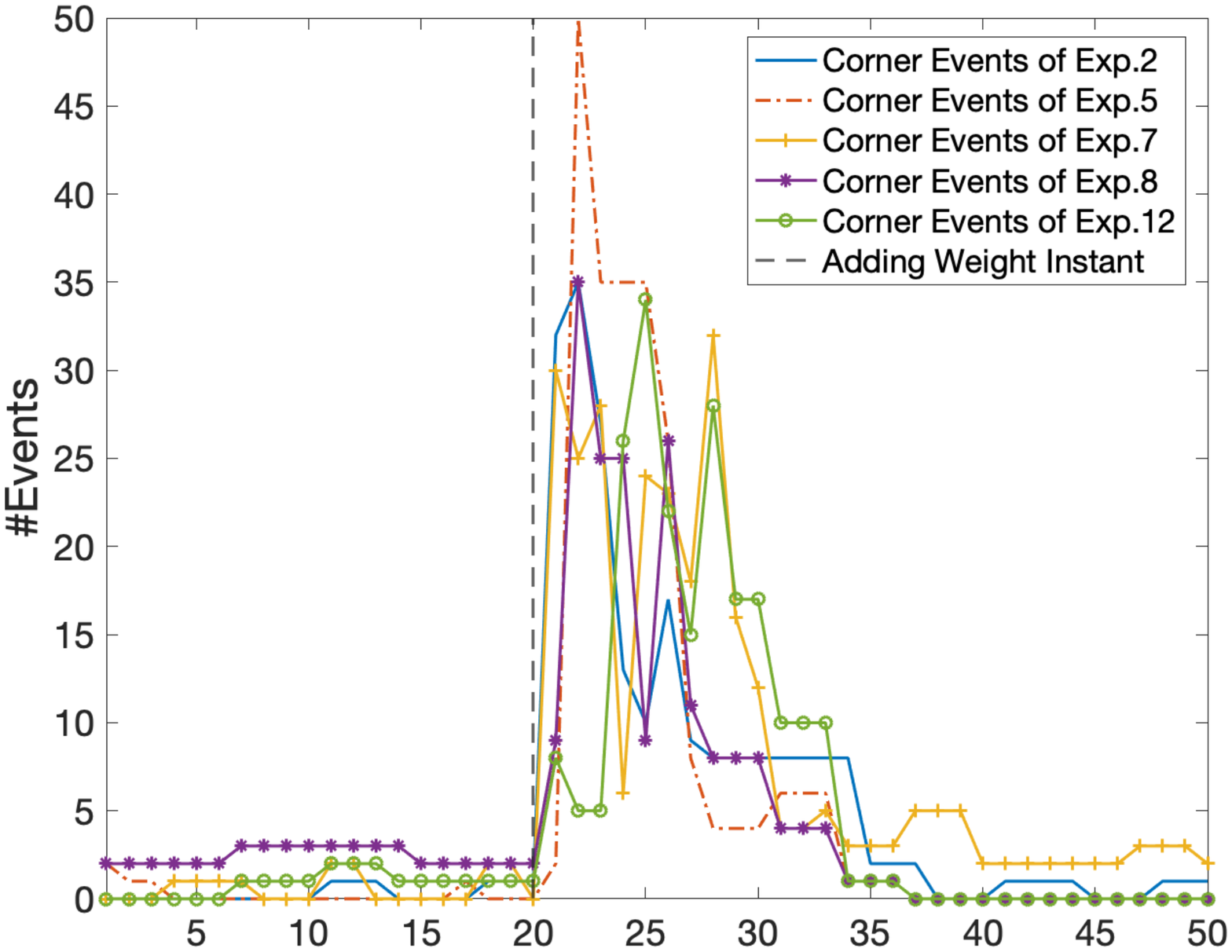}}
\caption{Light weight case: Slip patterns of feature and raw events analysed from five experiments}
\label{slip_patterns}
\vspace{-1em}\end{figure}
\section{Slip Suppression Experiments and Results }

\begin{figure}[h!]
  \centering
      \subfloat[Fuzzy control applied for object slips detected under grasped condition]{\label{fig:sf1}
      \includegraphics[clip,trim = 1.5cm 0cm 0cm 1.5cm, keepaspectratio,width=0.5\textwidth, height=1\textwidth]{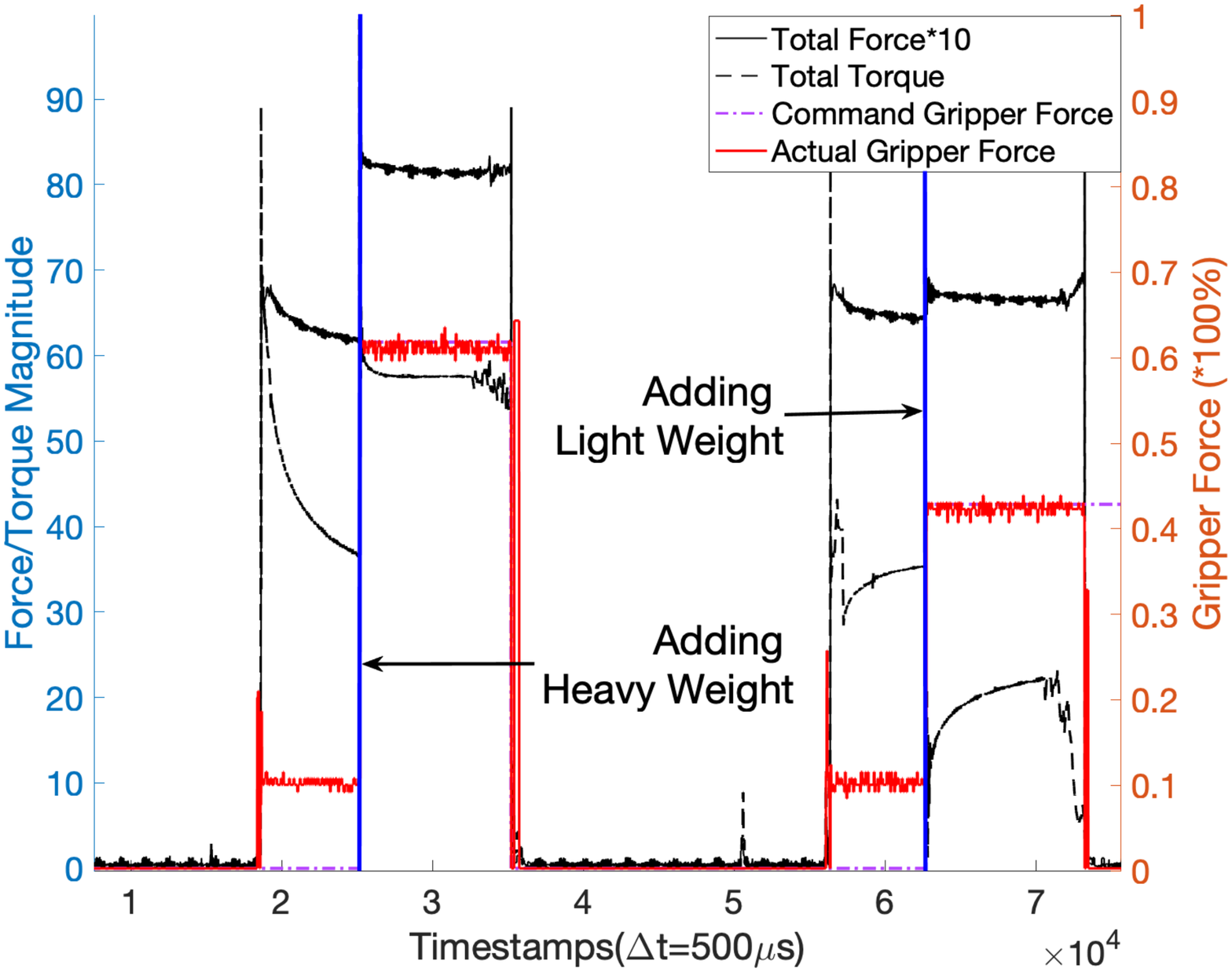}} 
\caption{Real-Time fuzzy grip force control for slips detected under varying loads}
\label{load_addition}\vspace{-1em}
\end{figure}
We use mamdani type fuzzy logic controller to adjust grasping force based on the incipient slip feedback such that the induced slip is suppressed. The rule base is setup with two inputs that is the number of edge and corner events detected at the initial time sample of a slip. Three triangluar membership function expressed in the linguistic variables as small(S), medium (M) and large (L) are chosen for the input variables and they are divided in to three equal parts. The range of the membership functions for the two inputs is chosen based on the min and max of the edge and corner events from several experimental trials. Five Gaussian membership function expressed in the linguistic variables as very small (VS), small(S), medium (M), large (L) and very large (VL) are chosen for the output variable and they are divided in to five equal parts. The range of output membership functions is selected according to the percent of holding force (0-100) applied by the gripper. The fuzzy rule base designed for the slip suppression strategy is given in Table \ref{slip_patterns}. 
\begin{table}
\centering
\caption{Fuzzy Control Logic}
\begin{tabular}{|p{0.6in}|C{0.3in}|C{0.4in}|C{0.4in}|C{0.4in}|} \hline 
 \multicolumn{2}{|c|}{\multirow{2}{*}{$Features$}}& \multicolumn{3}{c|}{Edge $ S_{e}$}   \\ \cline{3-5} 
 \multicolumn{2}{|c|}{}  & S & M & L \\ \hline 
  & S & VS & S & M \\ \cline{2-5} 
 Corner $S_{c}$ & M & S & M & L \\ \cline{2-5} 
 & L & M & L & VL \\ \hline 
\end{tabular} \label{slip_patterns}
\end{table}


\subsection{Slip suppression during Loading}
The addition of weight from a particular height generates different slips for each experiment. Thus, the slip patterns slightly varies for the same weight repeated experiments. moreover, the slip magnitude and pattern varies significantly for different weight additions. In Fig. \ref{slip_patterns}, in (a and b), the corner and edge stream of events indicating the slip patterns from five experiments for the addition of weight is presented. The mean and standard deviation of the featured events were calculated and used in setting the range of input membership functions. In Fig. \ref{load_addition} illustrates varying grip force for the induced slip caused by the addition of weight.


\begin{figure*}
  \centering
      \begin{tabular}{c c c c c c}            
      \subfloat{\label{fig:sf1}
      \includegraphics[width=0.17\textwidth, height=0.25\textwidth]{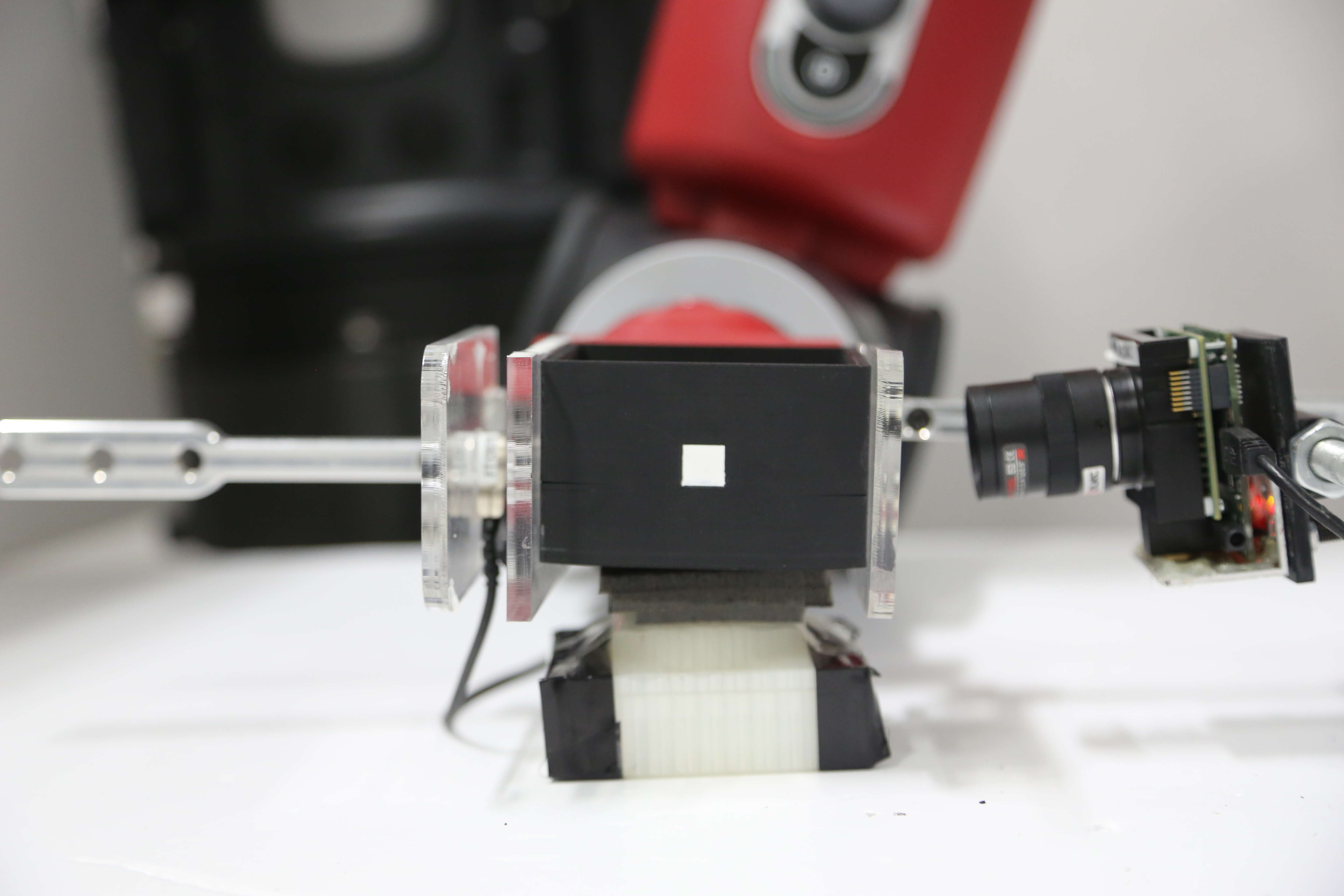}}  
\subfloat{\label{fig:sf1}
      \includegraphics[width=0.17\textwidth, height=0.25\textwidth]{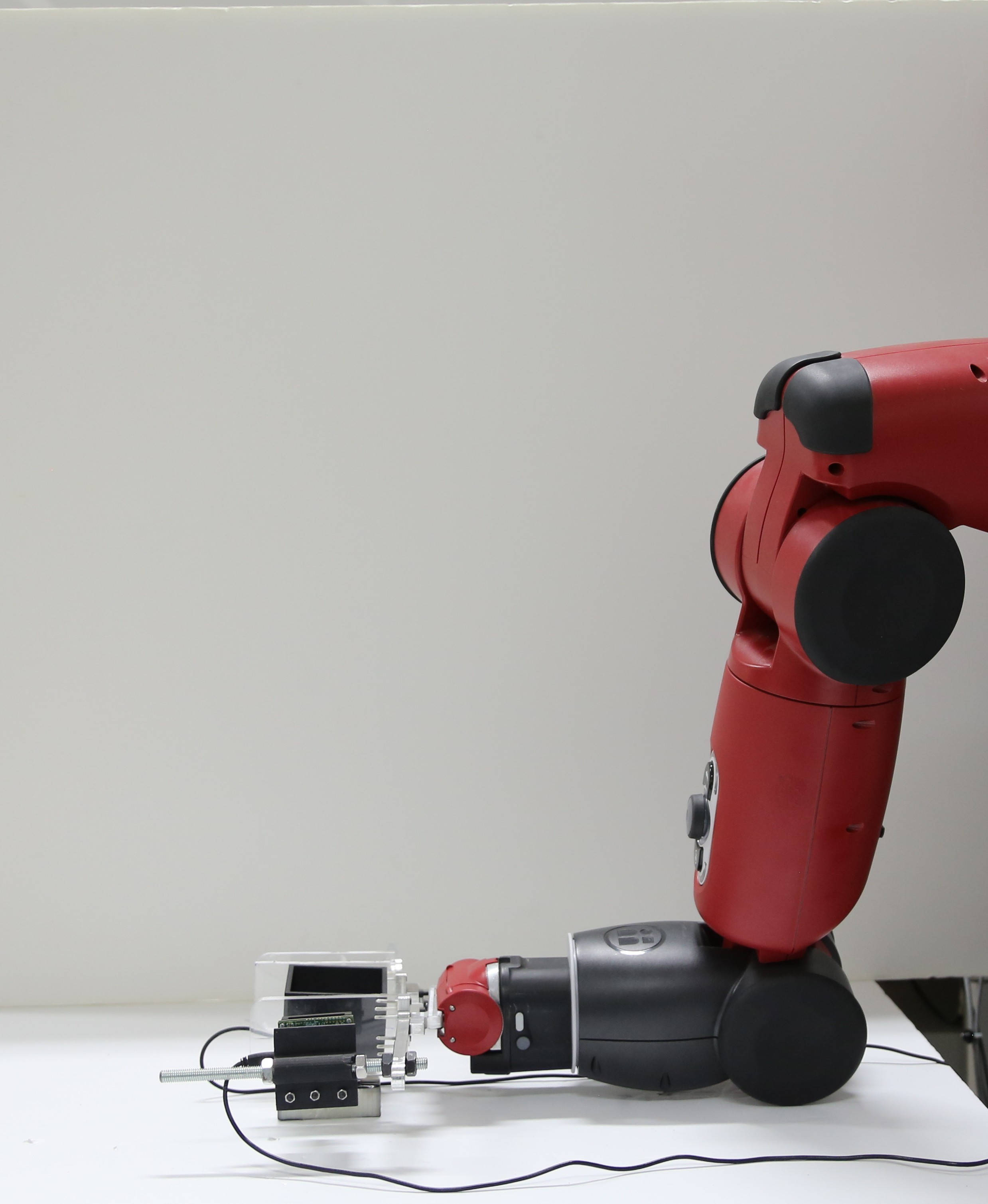}}  
      \subfloat{\label{fig:sf1}
      \includegraphics[width=0.21\textwidth, height=0.25\textwidth]{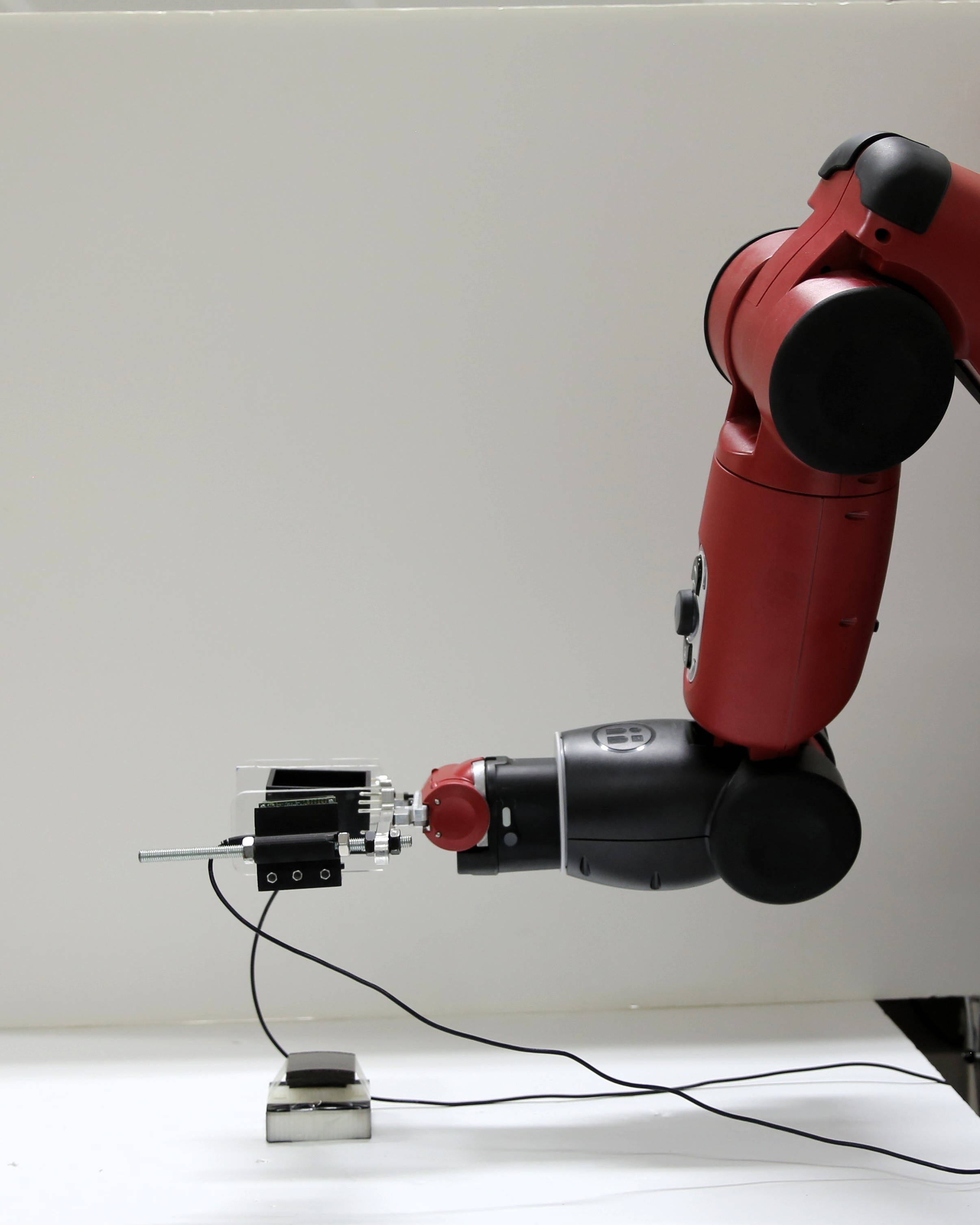}}  
      \subfloat{\label{fig:sf1}
      \includegraphics[width=0.2\textwidth, height=0.25\textwidth]{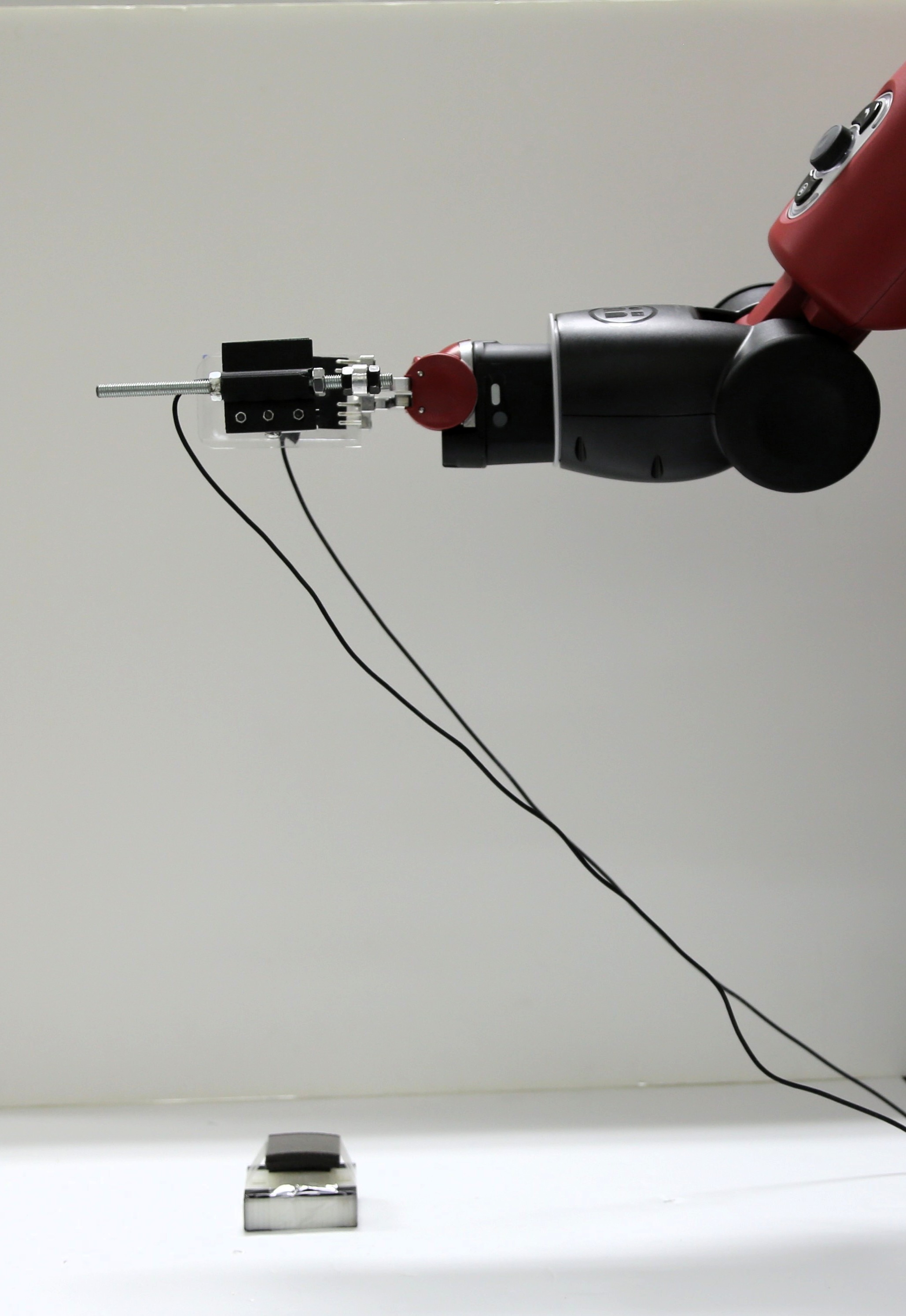}}  
        \subfloat{\label{fig:sf1}
      \includegraphics[width=0.18\textwidth, height=0.25\textwidth]{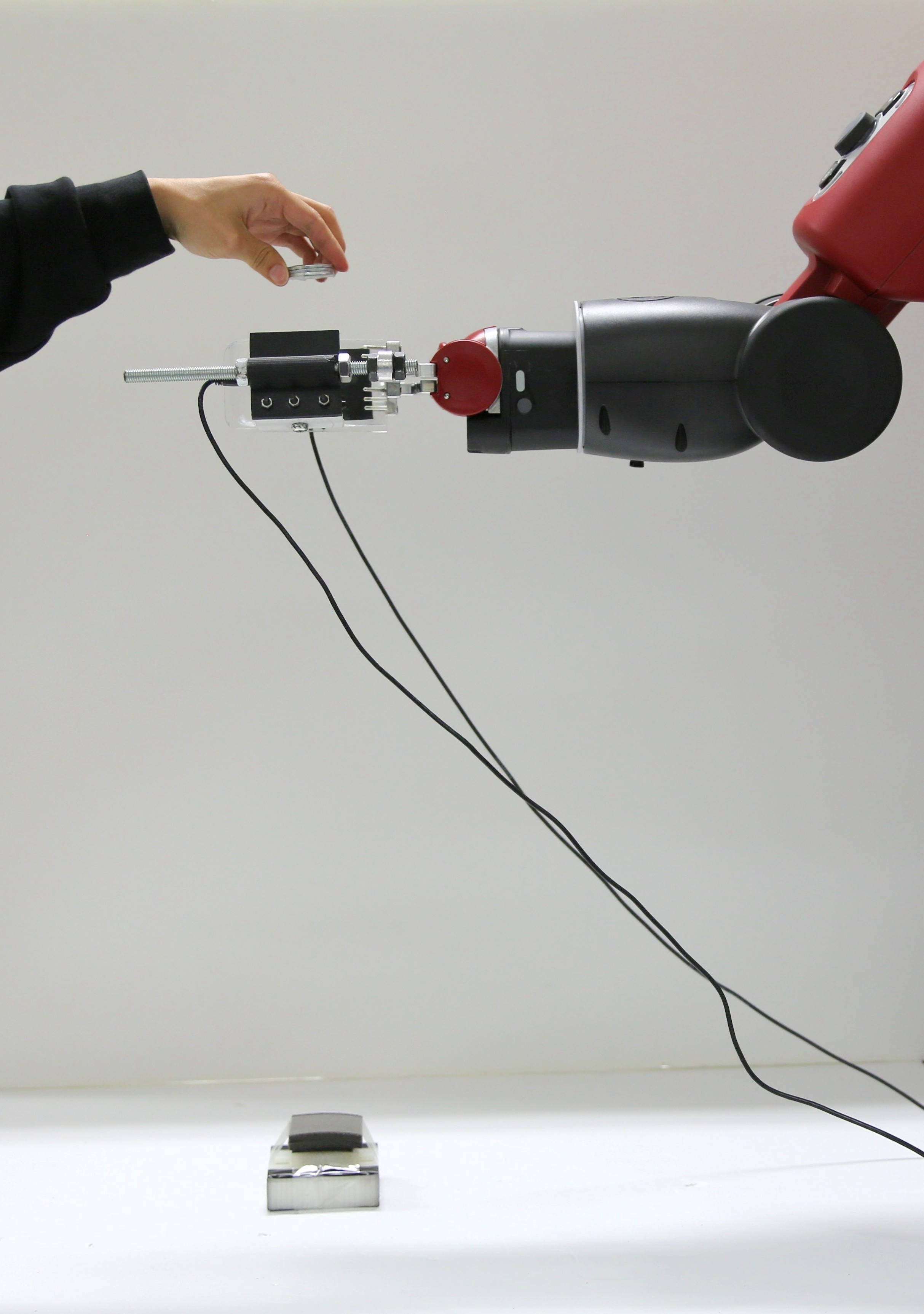}}
      \end{tabular}      \\
      \vspace{-7mm}
       \subfloat{            \adjustbox{left}{

      \includegraphics[clip,trim = 2.5cm 1.5cm 0cm 0.5cm, keepaspectratio,width=1.09\textwidth, height=1.2\textwidth]{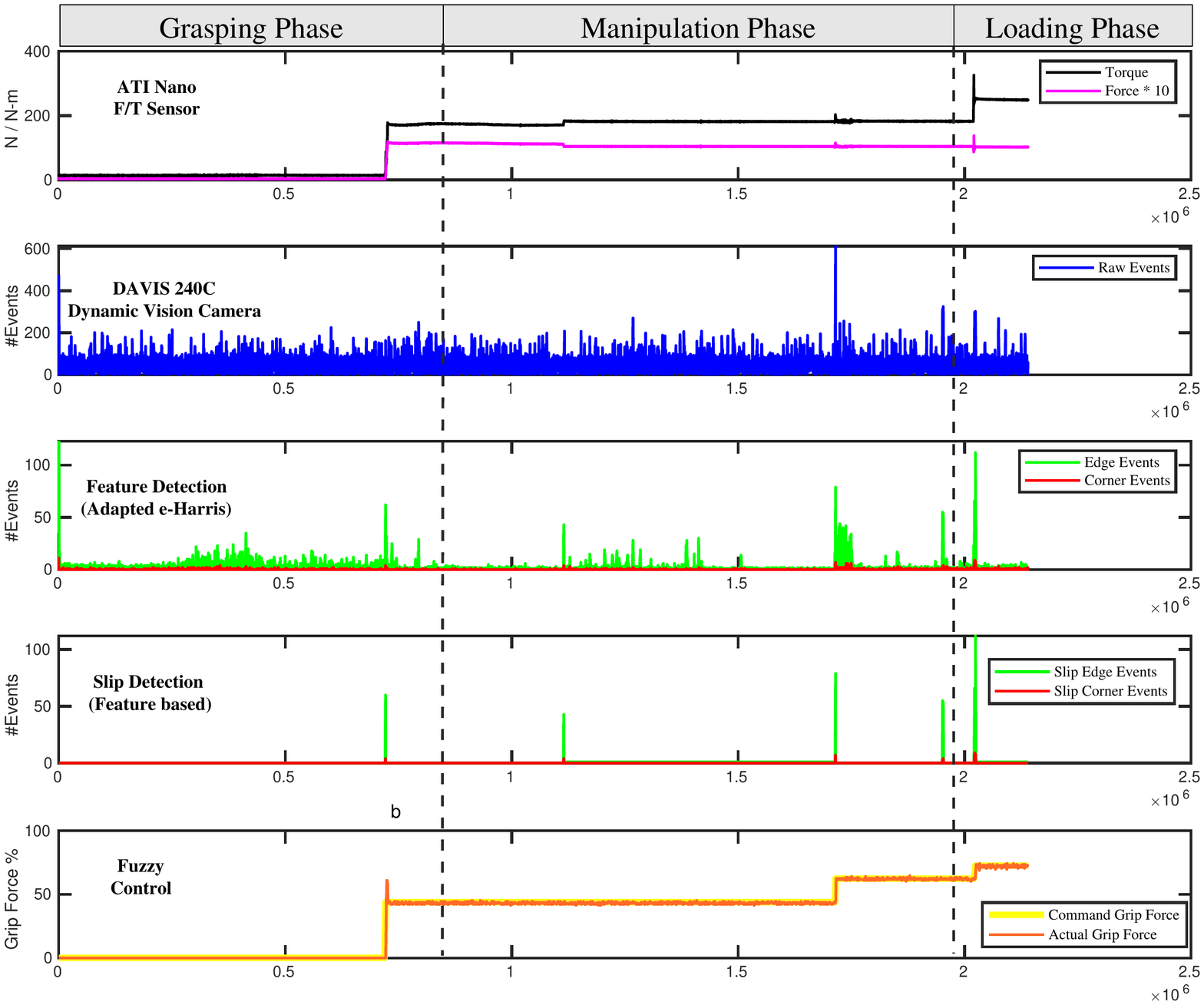}}}
\caption{Demonstration of slip detection and suppression in robotic object manipulation: (a) Grasping phase:  Object is caged first; while lifting initially, slip caused primarily by the insufficient grip force is detected and tackled with 43 \% grip force; Manipulation Phase: slips caused by the Dynamic movement of the manipulator is detected and only the significant slip is tackled with 63 \% grip force; Loading phase: slip caused by the addition of weight is tackled with 72 \% grip force.   }
\label{complete_manipulation}
\end{figure*}

\begin{figure*}
  \centering
      \begin{tabular}{c c c c c c}            
      \subfloat{\label{fig:sf1}
      \includegraphics[width=0.23\textwidth, height=0.25\textwidth]{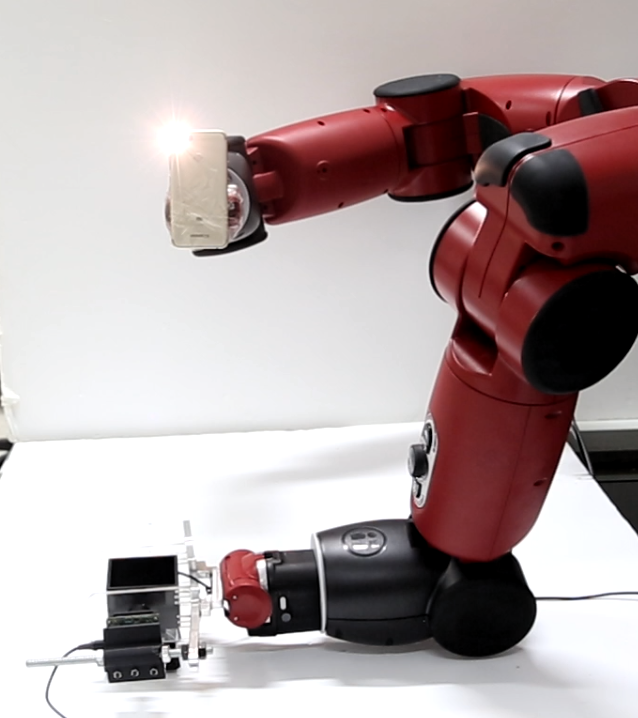}}
            \subfloat{\label{fig:sf1}
      \includegraphics[width=0.12\textwidth, height=0.25\textwidth]{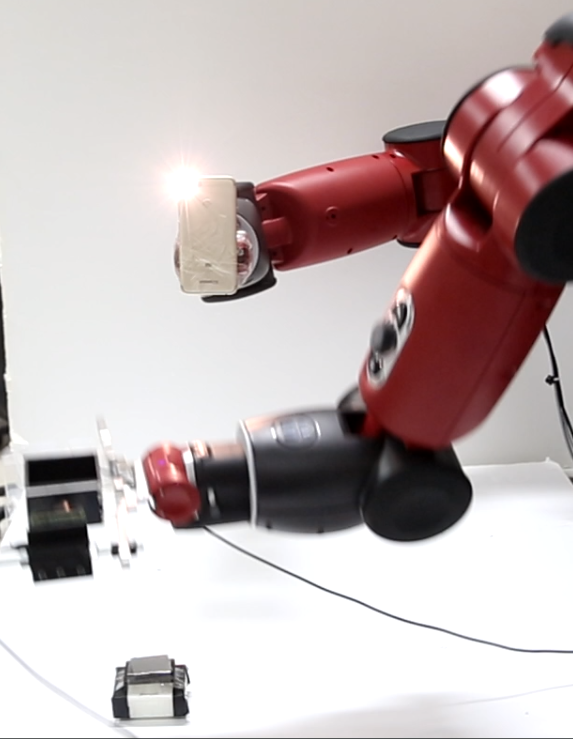}}   
\subfloat{\label{fig:sf1}
      \includegraphics[width=0.12\textwidth, height=0.25\textwidth]{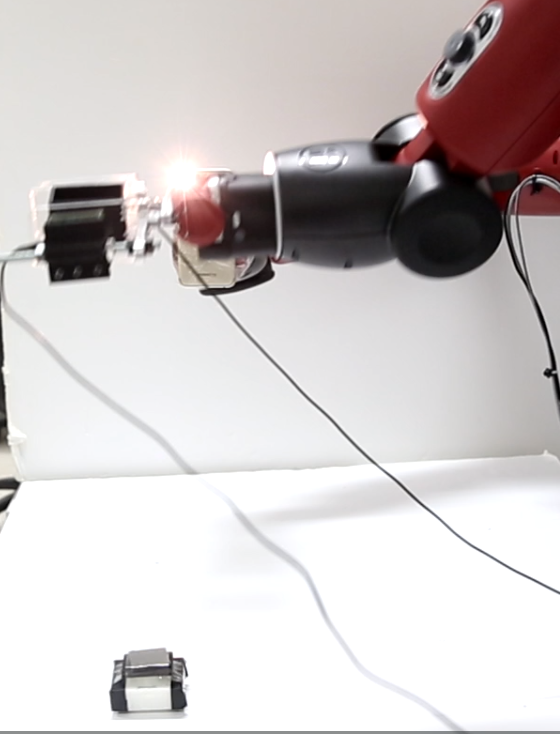}}  
      \subfloat{\label{fig:sf1}
      \includegraphics[width=0.20\textwidth, height=0.25\textwidth]{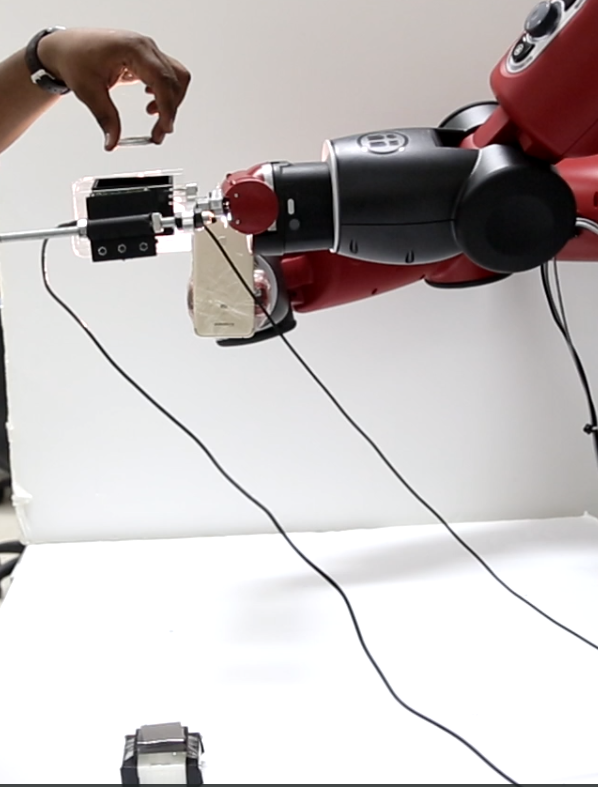}}  
      \subfloat{\label{fig:sf1}
      \includegraphics[width=0.14\textwidth, height=0.25\textwidth]{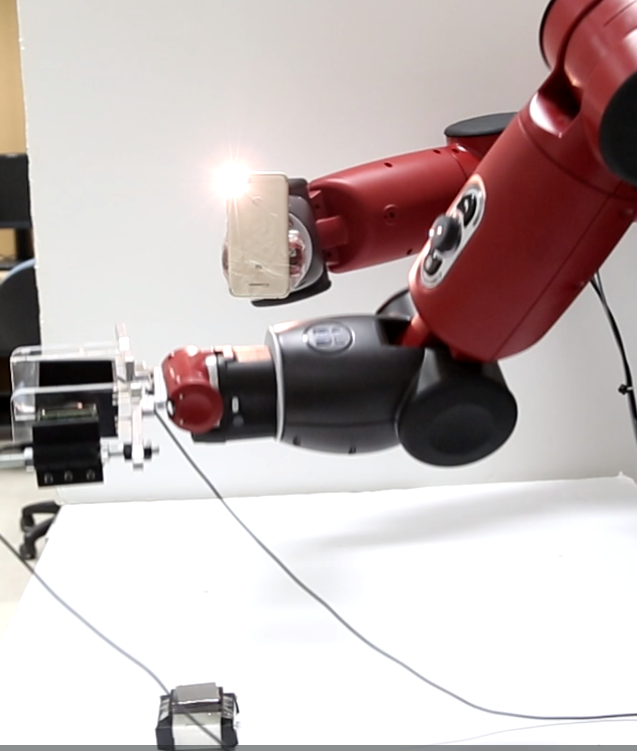}}  
        \subfloat{\label{fig:sf1}
      \includegraphics[width=0.14\textwidth, height=0.25\textwidth]{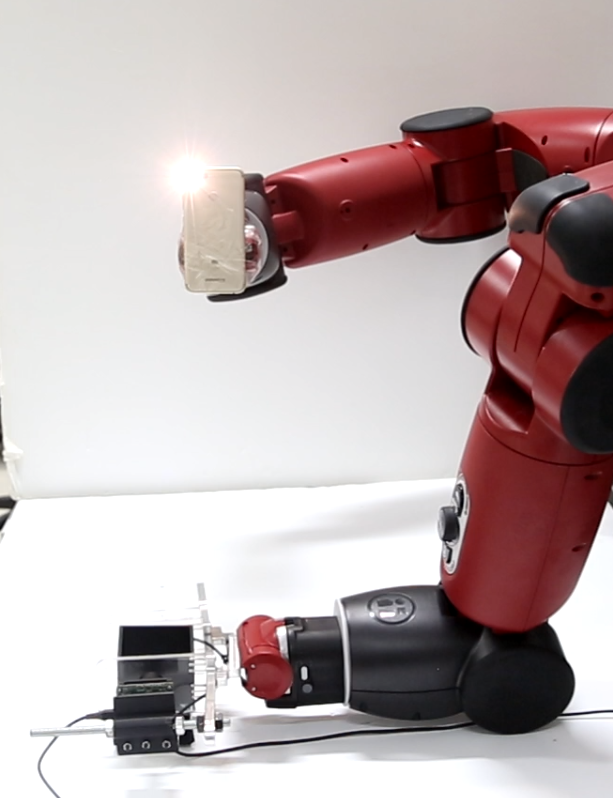}}
      \end{tabular}      \\
      \vspace{-7mm}
       \subfloat{            \adjustbox{left}{

      \includegraphics[clip,trim = 2.5cm 1.5cm 0cm 0.5cm, keepaspectratio,width=1.09\textwidth, height=1.2\textwidth]{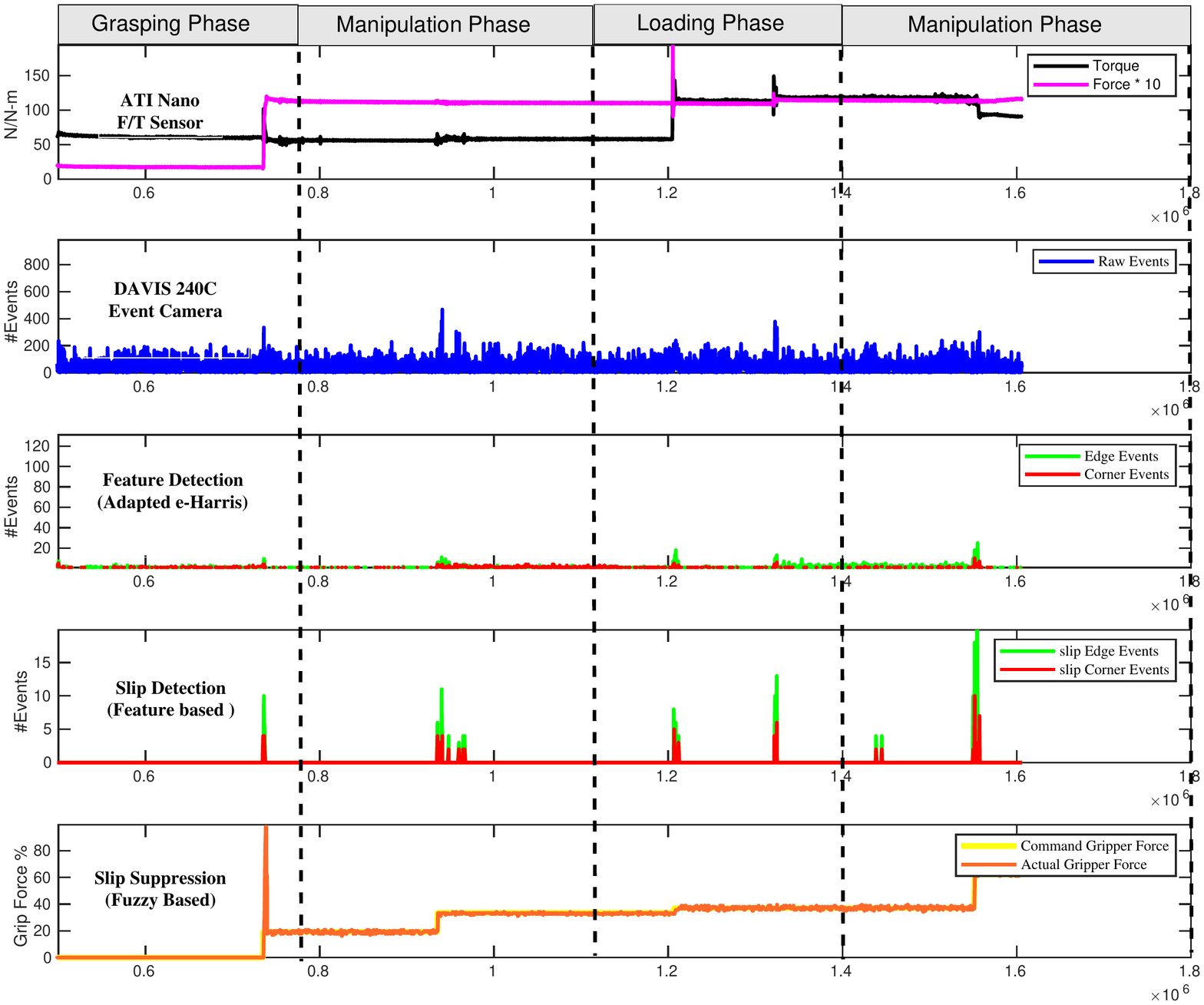}}}
\caption{Demonstration of slip detection and suppression in robotic object manipulation under illumination and vibration uncertainty: (a) Grasping phase:  Object is caged first; while lifting initially, slip caused primarily by the insufficient grip force is detected and tackled with 17 \% grip force; Manipulation Phase: slips caused by the Dynamic movement of the manipulator is detected and only the significant slip is tackled with 33 \% grip force; Loading phase: slip caused by the addition of weight is tackled with 37 \% grip force; Manipulation Phase: slips caused by the Dynamic movement of the manipulator while lowering and surface contact is detected and only the significant slip is tackled with 62 \% grip force;.   }
\label{complete_manipulation_uncertainty}
\end{figure*}
\subsection{Slip Detection and Suppression during Robotic Grasping and Object Manipulation}


Fig. \ref{complete_manipulation} and \ref{complete_manipulation_uncertainty} demonstrates the grasping, manipulation and loading phases of robotic object manipulation with the corresponding images and signals from F/T sensor, event camera, feature detector, slip detector and fuzzy controller at a sampling rate of $500  \mu s$. The first row depicts the sequence of operation in order and aligning to the phases of the experiment. F/T changes in the second row reflect the induced object slip and force adjustments made by the gripper. Feature detector that detects the edges and corner from the sampled event stream (row three) is plotted in the fourth row. Slips detected by the feature-based approach and grip force commanded by the fuzzy controller to suppress slips is illustrated in the last two rows. The proposed suppression strategy allows only the improvements of the grip force starting from the initial grasp to the end of manipulation. 

Fig. \ref{complete_manipulation} illustrates slip detection and suppression performance under vibration uncertainty. In the grasping phase, the robot manipulator reaches the pre-grasp pose planned for the known box object and the object is caged. Then, the manipulator slightly does a upward motion (0.05 m) to induce slip under caged condition such that a sufficient force is applied while lifting. Based on the incipient slip detected, the fuzzy controller determine a grip force to hold the box against gravity. In this experiment, a 43 \% of grip force is applied to suppress the slip detected during such grasp adjustments. The actual grip force overshoots due to the initial grasp of the object and later settle back to the commanded grip force.

In the manipulation phase, the robot left end effector is commanded to move straight up at a height of 0.45 meter at a constant speed of 1 m/s. During the manipulator motion, both internal and external disturbances causes object slips. In the experiment, three slip instances detected and the slip which causes instability to the object is tackled by the fuzzy controller with a grip force of 63 \%.

In the loading phase, a human user drops the load (200 gm) from a height ranging between 4 cm to 8 cm above the grasped object. The feature based approach detects slip with a highest count of corners and edges from the overall phases and a grip force of 72 \% was applied to suppress the slip. We performed five experiments, each following the same procedure above and covering all the phases of robotic object manipulation. The grip force slightly varied in all phases of the experiment due to the detected slip variations caused by object pose uncertainty.

 Fig. \ref{complete_manipulation_uncertainty} illustrates slip detection and suppression performance under both illumination and vibration uncertainty. The right end effector of the Baxter robot with an embedded light source is positioned at a height of 0.45 m from the table and maintains a distance of 20 cm with respect to the left arm. Apart from this setting, the experiment is conducted in a similar fashion, described above for Fig. \ref{complete_manipulation}. In this experiment, a 17 \% of grip force is applied to suppress the slip detected during initial grasp. A large overshoot occurred due to the small grip force selected. During the object manipulation task, the varying illumination generates more noisy spikes in the event camera. Even-though our robust approach tackled such illumination uncertainty, few false slips got detected. However, our controller operates in an incremental manner and reacts to only incipient slips that is greater in magnitude to the earlier ones. We observed the manipulation phase during lifting and lowering operations, where the incipient slips that can affect the stability of the object grasp is tackled with 33 \% and 62 \% of grip force. In the loading phase, load addition is tackled with a 37 \% grip force.  In repeated experiments, we varied the position of light source mounted end effector to rigorously test our slip detector under illumination uncertainty.

To evaluate the performance of our even-based method, we propose a slip metric ($Q_{sm}$) that quantifies the object position deviation under grasped condition. Two event-based frames are captured that holds the movement of the marker after the initial grasp and completion of manipulation task. In both event frames, the centroid of the marker $  M^{c} = \bar{x_{i}}, \bar{y_{i}} =(\sum x_{i}/n, \sum y_{i}/n )$ is computed from a set of $n$ detected corner points. The slip metric is nothing but a euclidean distance between two centroid points expressed as
\begin{equation}
Q_{sm} = d(M_{s}^{c},M_{f}^{c})
\end{equation}
where $M_{s}^{c}$ and $M_{f}^{c}$ represents the centroid of the marker computed after the initial grasp and before the grasp release. We consider this position error as our slip metric to quantify the ability of the approach that enforce to preserve grasp stability. Performance measures observed in multiple experiments are presented in Table. \ref{metric} where the metric and measures from five trials under vibration and illumination uncertainty is given in (a) and (b) accordingly. $N_{control}$ indicates the number of force adjustments made corresponding to feature based slip detector in the overall task.


The average of slip measures from three trials in (a) and (b) is 2.12 mm and 2.5 mm. Besides, a low standard deviation is indicated in both cases. This emphasize the ability of the event-based method that considers both slip detection and suppression to achieve high precision in manipulation task. These errors are mainly attributed to design imperfections such as the misalignment of the parallel fingers and the limitations of the gripper motor to respond quickly. The precision performance can be improved by better gripper design and motor ability to reach a certain force/torque in a microsecond level.

Our proposed event-based finger-vision system can handle objects (small) that can be seen within the limits of the finger boundaries and objects (large) with textures. Our proposed method enables the slip detection approaches to  autonomously set a threshold in real-time without requiring any object knowledge such that our robust slip detection approach is generic. However, the proposed control strategy requires knowledge of the slip events occured under grasped condition from experimental trials. In our future work, we would like to devise suppression strategy that is generic while incorporating bio-inspired models.

 To test the applicability of the system and approach in a standard form, markers of different primitive shapes (circle, rectangle and square) carved on the sides of the test object were used in the experiments. We observed satisfactory performances in detecting slips and suppressing them during object manipulation. Overall, the feature based slip detection and fuzzy logic based suppression stratergy achieved a 100 \% success rate by avoiding grasp failures and shown superior performance in maintaining stable object grasp in all experiments under vibration and illumination uncertainty.

\begin{table}[t]
\centering
\subfloat[Under vibration uncertainty]
 {
\centering
\quad \quad
\begin{tabular}{c|cc}
\hline
\multicolumn{1}{l|}{Trial} & \multicolumn{1}{l}{$N_{control}$}  & \multicolumn{1}{l}{$Q_{sm} (mm)$} \\ \hline
1                          & 2                 & 3.6                                     \\
2                          & 3                               & 1.4                                     \\
3                          & 3                             & 1.8                                     \\ 
4 (Fig. \ref{complete_manipulation})                         & 3                                                      & 2.1                                     \\ 
5                          & 3                                                     & 1.7                                     \\ \hline
\end{tabular}
}

\qquad

\subfloat[Under vibration and illumination uncertainty]
{
\centering
\quad \quad
\begin{tabular}{c|cccc}
\hline
\multicolumn{1}{l|}{Trial} & \multicolumn{1}{l}{$N_{control}$} & \multicolumn{1}{l}{$Q_{sm} (mm)$} \\ \hline
1                          & 3                                                       & 3.4                                     \\
2                          & 3                                                      & 1.2                                     \\
3                          & 4                                               & 3.7                                     \\ 
4                          & 4                                                   & 1.4                                     \\ 
5 (Fig. \ref{complete_manipulation_uncertainty})                         & 4                                                      & 2.7                                     \\ \hline
\end{tabular}
}
\caption{Performance measures of Event-based method: Slip detection and suppression in robotic grasping and manipulation.}
\label{metric}\vspace{-1em}
\end{table}

\section{Conclusion}

Event cameras are biomimetic vision sensors having fundamentally different sensing mechanics to conventional sensors. We presented event-based finger vision system for robotic grippers with simple settings to tackle slip incidents starting from object grasping till manipulation task completion. An online method that involves noise sampling for calibration, grasping, slip detection and suppression for maintaining grasp stability is introduced. Two approaches, a baseline and feature based approach for event-based slip detection and a mamdani-type fuzzy controller to adjust the gripping force using incipient slip feedback were proposed. The performances of the approaches were studied under two sampling rates, different noise levels and three state of the art corner detectors. 

The feature based approach detected incipient slip at a sampling rate of 2kHz, gave higher accuracy over baseline approach and proven robust to illumination and vibrations uncertainties. Average slip metric values obtained from complete robotic object manipulation experiments validated high-performance precision manipulation. The timely detection of slips and intelligent grasp force adjustments to suppress slip demonstrated in experiments emphasize their superiority over traditional tactile sensing and applicability in industrial-grade robotic automation. 



In our future work, we would like to develop marker free event-based slip detection approaches to handle large and textured objects. Moreover, we would like to equip event-based finger vision system with multi-modal functionality to sense force distribution, object pose and texture; develop neuromorphic vision based grippers; investigate event learning-based methods to detect and suppress object slips. Devising purely event based slip detection and suppression strategies to handle dynamic slip at any point during grasping and manipulation is an interesting area to explore.



\section*{Acknowledgment}

This work is supported by the Khalifa University of Science and Technology under Award No. CIRA-2018-55 and RC1-2018-KUCARS.
\vspace{-1em}
\ifCLASSOPTIONcaptionsoff
  \newpage
\fi

\bibliographystyle{IEEEtran}
\bibliography{Robust_slip_detection_frame_based_2019} 

\begin{thebibliography}{10}
\providecommand{\url}[1]{#1}
\csname url@samestyle\endcsname
\providecommand{\newblock}{\relax}
\providecommand{\bibinfo}[2]{#2}
\providecommand{\BIBentrySTDinterwordspacing}{\spaceskip=0pt\relax}
\providecommand{\BIBentryALTinterwordstretchfactor}{4}
\providecommand{\BIBentryALTinterwordspacing}{\spaceskip=\fontdimen2\font plus
\BIBentryALTinterwordstretchfactor\fontdimen3\font minus
  \fontdimen4\font\relax}
\providecommand{\BIBforeignlanguage}[2]{{%
\expandafter\ifx\csname l@#1\endcsname\relax
\typeout{** WARNING: IEEEtran.bst: No hyphenation pattern has been}%
\typeout{** loaded for the language `#1'. Using the pattern for}%
\typeout{** the default language instead.}%
\else
\language=\csname l@#1\endcsname
\fi
#2}}
\providecommand{\BIBdecl}{\relax}
\BIBdecl

\bibitem{aceto2019survey}
G.~Aceto, V.~Persico, and A.~Pescap{\'e}, ``A survey on information and
  communication technologies for industry 4.0: State-of-the-art, taxonomies,
  perspectives, and challenges,'' \emph{IEEE Communications Surveys \&
  Tutorials}, vol.~21, no.~4, pp. 3467--3501, 2019.

\bibitem{de2019grasping}
P.~De~La~Puente, D.~Fischinger, M.~Bajones, D.~Wolf, and M.~Vincze, ``Grasping
  objects from the floor in assistive robotics: Real world implications and
  lessons learned,'' \emph{IEEE Access}, vol.~7, pp. 123\,725--123\,735, 2019.

\bibitem{birglen2018statistical}
L.~Birglen and T.~Schlicht, ``A statistical review of industrial robotic
  grippers,'' \emph{Robotics and Computer-Integrated Manufacturing}, vol.~49,
  pp. 88--97, 2018.

\bibitem{chen2018tactile}
W.~Chen, H.~Khamis, I.~Birznieks, N.~F. Lepora, and S.~J. Redmond, ``Tactile
  sensors for friction estimation and incipient slip detection—toward
  dexterous robotic manipulation: A review,'' \emph{IEEE Sensors Journal},
  vol.~18, no.~22, pp. 9049--9064, 2018.

\bibitem{li2019survey}
R.~Li and H.~Qiao, ``A survey of methods and strategies for high-precision
  robotic grasping and assembly tasks—some new trends,'' \emph{IEEE/ASME
  Transactions on Mechatronics}, vol.~24, no.~6, pp. 2718--2732, 2019.

\bibitem{vanarse2016review}
A.~Vanarse, A.~Osseiran, and A.~Rassau, ``A review of current neuromorphic
  approaches for vision, auditory, and olfactory sensors,'' \emph{Frontiers in
  neuroscience}, vol.~10, p. 115, 2016.

\bibitem{liu2010neuromorphic}
S.-C. Liu and T.~Delbruck, ``Neuromorphic sensory systems,'' \emph{Current
  opinion in neurobiology}, vol.~20, no.~3, pp. 288--295, 2010.

\bibitem{yole2019}
\BIBentryALTinterwordspacing
{Yole Développement}, ``Neuromorphic sensing and computing,'' 2019, [Online;
  2019]. [Online]. Available:
  \url{https://yole-i-micronews-com.osu.eu-west-2.outscale.com/uploads/2019/09/YD19039_Neuromorphic_Sensing_Computing2019_sample.pdf}
\BIBentrySTDinterwordspacing

\bibitem{indiveri2000neuromorphic}
G.~Indiveri and R.~Douglas, ``Neuromorphic vision sensors,'' \emph{Science},
  vol. 288, no. 5469, pp. 1189--1190, 2000.

\bibitem{ieng2014asynchronous}
S.-H. Ieng, C.~Posch, and R.~Benosman, ``Asynchronous neuromorphic event-driven
  image filtering,'' \emph{Proceedings of the IEEE}, vol. 102, no.~10, pp.
  1485--1499, 2014.

\bibitem{gollisch2010eye}
T.~Gollisch and M.~Meister, ``Eye smarter than scientists believed: neural
  computations in circuits of the retina,'' \emph{Neuron}, vol.~65, no.~2, pp.
  150--164, 2010.

\bibitem{bensmaia2008representation}
S.~J. Bensmaia, P.~V. Denchev, J.~F. Dammann, J.~C. Craig, and S.~S. Hsiao,
  ``The representation of stimulus orientation in the early stages of
  somatosensory processing,'' \emph{Journal of Neuroscience}, vol.~28, no.~3,
  pp. 776--786, 2008.

\bibitem{yau2009analogous}
J.~M. Yau, A.~Pasupathy, P.~J. Fitzgerald, S.~S. Hsiao, and C.~E. Connor,
  ``Analogous intermediate shape coding in vision and touch,''
  \emph{Proceedings of the National Academy of Sciences}, vol. 106, no.~38, pp.
  16\,457--16\,462, 2009.

\bibitem{pruszynski2014edge}
J.~A. Pruszynski and R.~S. Johansson, ``Edge-orientation processing in
  first-order tactile neurons,'' \emph{Nature neuroscience}, vol.~17, no.~10,
  p. 1404, 2014.

\bibitem{venkataramani2010orientation}
S.~Venkataramani and W.~R. Taylor, ``Orientation selectivity in rabbit retinal
  ganglion cells is mediated by presynaptic inhibition,'' \emph{Journal of
  Neuroscience}, vol.~30, no.~46, pp. 15\,664--15\,676, 2010.

\bibitem{iancu2012mamdani}
I.~Iancu, ``A mamdani type fuzzy logic controller,'' in \emph{Fuzzy
  logic-controls, concepts, theories and applications}.\hskip 1em plus 0.5em
  minus 0.4em\relax IntechOpen, 2012.

\bibitem{dahiya2009tactile}
R.~S. Dahiya, G.~Metta, M.~Valle, and G.~Sandini, ``Tactile sensing—from
  humans to humanoids,'' \emph{IEEE transactions on robotics}, vol.~26, no.~1,
  pp. 1--20, 2009.

\bibitem{maher1989implementing}
M.~A.~C. Maher, S.~P. Deweerth, M.~A. Mahowald, and C.~A. Mead, ``Implementing
  neural architectures using analog vlsi circuits,'' \emph{IEEE transactions on
  circuits and systems}, vol.~36, no.~5, pp. 643--652, 1989.

\bibitem{vallbo1984properties}
A.~B. Vallbo, R.~S. Johansson \emph{et~al.}, ``Properties of cutaneous
  mechanoreceptors in the human hand related to touch sensation,'' \emph{Hum
  Neurobiol}, vol.~3, no.~1, pp. 3--14, 1984.

\bibitem{westling1987responses}
G.~Westling and R.~S. Johansson, ``Responses in glabrous skin mechanoreceptors
  during precision grip in humans,'' \emph{Experimental brain research},
  vol.~66, no.~1, pp. 128--140, 1987.

\bibitem{romano2011human}
J.~M. Romano, K.~Hsiao, G.~Niemeyer, S.~Chitta, and K.~J. Kuchenbecker,
  ``Human-inspired robotic grasp control with tactile sensing,'' \emph{IEEE
  Transactions on Robotics}, vol.~27, no.~6, pp. 1067--1079, 2011.

\bibitem{nakagawa2019bio}
A.~Nakagawa-Silva, N.~V. Thakor, J.-J. Cabibihan, and A.~B. Soares, ``A
  bio-inspired slip detection and reflex-like suppression method for robotic
  manipulators,'' \emph{IEEE Sensors Journal}, vol.~19, no.~24, pp.
  12\,443--12\,453, 2019.

\bibitem{jung2011biohybrid}
R.~Jung, \emph{Biohybrid systems: nerves, interfaces and machines}.\hskip 1em
  plus 0.5em minus 0.4em\relax John Wiley \& Sons, 2011.

\bibitem{gollisch2009throwing}
T.~Gollisch, ``Throwing a glance at the neural code: rapid information
  transmission in the visual system,'' \emph{HFSP journal}, vol.~3, no.~1, pp.
  36--46, 2009.

\bibitem{lichtsteiner2008128}
P.~Lichtsteiner, C.~Posch, and T.~Delbruck, ``Latency asynchronous temporal
  contrast vision sensor,'' \emph{IEEE journal of solid-state circuits},
  vol.~43, no.~2, pp. 566--576, 2008.

\bibitem{posch2010qvga}
C.~Posch, D.~Matolin, and R.~Wohlgenannt, ``A qvga 143 db dynamic range
  frame-free pwm image sensor with lossless pixel-level video compression and
  time-domain cds,'' \emph{IEEE Journal of Solid-State Circuits}, vol.~46,
  no.~1, pp. 259--275, 2010.

\bibitem{berner2013240}
R.~Berner, C.~Brandli, M.~Yang, S.-C. Liu, and T.~Delbruck, ``latency
  sparse-output vision sensor for mobile applications,'' in \emph{2013
  Symposium on VLSI Circuits}.\hskip 1em plus 0.5em minus 0.4em\relax IEEE,
  2013, pp. C186--C187.

\bibitem{brandli2014240}
C.~Brandli, R.~Berner, M.~Yang, S.-C. Liu, and T.~Delbruck, ``latency global
  shutter spatiotemporal vision sensor,'' \emph{IEEE Journal of Solid-State
  Circuits}, vol.~49, no.~10, pp. 2333--2341, 2014.

\bibitem{Anne2019}
\BIBentryALTinterwordspacing
{Anne-Françoise Pelé }, ``Event-driven vision hits production lines,'' 2019,
  [Online; 2019]. [Online]. Available:
  \url{https://www.eetimes.com/event-driven-vision-hits-production-lines/#}
\BIBentrySTDinterwordspacing

\bibitem{Bohg2014}
J.~Bohg, A.~Morales, T.~Asfour, and D.~Kragic, ``Data-driven grasp synthesis;a
  survey,'' \emph{IEEE Transactions on Robotics}, vol.~30, pp. 289--309, 2014.

\bibitem{LiSastry1988}
Z.~Li and S.~S. Sastry, ``Task-oriented optimal grasping by multifingered robot
  hands,'' \emph{IEEE Journal on Robotics and Automation}, vol.~4, pp. 32--44,
  1988.

\bibitem{LinSun2015}
Y.~Lin and Y.~Sun, ``Grasp planning to maximize task coverage,'' \emph{The
  International Journal of Robotics Research}, vol.~34, no.~9, pp. 1195--1210,
  2015.

\bibitem{salisbury1983kinematic}
J.~K. Salisbury and B.~Roth, ``Kinematic and force analysis of articulated
  mechanical hands,'' \emph{Journal of Mechanisms, Transmissions, and
  Automation in Design}, vol. 105, no.~1, pp. 35--41, 1983.

\bibitem{Ferrari1992}
C.~Ferrari and C.~John, ``Planning optimal grasps,'' in \emph{Proc. of the IEEE
  Int. Conf. on Robot. Autom.}, 1992, pp. 2290--2295.

\bibitem{Roa2015}
M.~A. Roa and R.~Suarez, ``Grasp quality measures: review and performance,''
  \emph{Autonomous Robots}, vol.~38, pp. 65--88, 2015.

\bibitem{melchiorri2000slip}
C.~Melchiorri, ``Slip detection and control using tactile and force sensors,''
  \emph{IEEE/ASME transactions on mechatronics}, vol.~5, no.~3, pp. 235--243,
  2000.

\bibitem{gunji2007grasping}
D.~Gunji, T.~Araki, A.~Namiki, A.~Ming, and M.~Shimojo, ``Grasping force
  control of multi-fingered robot hand based on slip detection using tactile
  sensor,'' \emph{Journal of the Robotics Society of Japan}, vol.~25, no.~6,
  pp. 970--978, 2007.

\bibitem{mizoguchi2010development}
Y.~Mizoguchi, K.~Tadakuma, H.~Hasegawa, A.~Ming, M.~Ishikawa, and M.~Shimojo,
  ``Development of intelligent robot hand using proximity, contact and slip
  sensing,'' \emph{Transactions of the Society of Instrument and Control
  Engineers}, vol.~46, no.~10, pp. 632--640, 2010.

\bibitem{shinoda2000instantaneous}
H.~Shinoda, S.~Sasaki, and K.~Nakamura, ``Instantaneous evaluation of friction
  based on artc tactile sensor,'' in \emph{Proceedings 2000 ICRA. Millennium
  Conference. IEEE International Conference on Robotics and Automation.
  Symposia Proceedings (Cat. No. 00CH37065)}, vol.~3.\hskip 1em plus 0.5em
  minus 0.4em\relax IEEE, 2000, pp. 2173--2178.

\bibitem{cotton2007novel}
D.~P. Cotton, P.~H. Chappell, A.~Cranny, N.~M. White, and S.~P. Beeby, ``A
  novel thick-film piezoelectric slip sensor for a prosthetic hand,''
  \emph{IEEE sensors journal}, vol.~7, no.~5, pp. 752--761, 2007.

\bibitem{dao2011development}
D.~V. Dao, S.~Sugiyama, S.~Hirai \emph{et~al.}, ``Development and analysis of a
  sliding tactile soft fingertip embedded with a microforce/moment sensor,''
  \emph{IEEE Transactions on Robotics}, vol.~27, no.~3, pp. 411--424, 2011.

\bibitem{dubey2006dynamic}
V.~N. Dubey and R.~M. Crowder, ``A dynamic tactile sensor on photoelastic
  effect,'' \emph{Sensors and Actuators A: Physical}, vol. 128, no.~2, pp.
  217--224, 2006.

\bibitem{roberts2011slip}
L.~Roberts, G.~Singhal, and R.~Kaliki, ``Slip detection and grip adjustment
  using optical tracking in prosthetic hands,'' in \emph{2011 Annual
  International Conference of the IEEE Engineering in Medicine and Biology
  Society}.\hskip 1em plus 0.5em minus 0.4em\relax IEEE, 2011, pp. 2929--2932.

\bibitem{tremblay1993estimating}
M.~R. Tremblay and M.~R. Cutkosky, ``Estimating friction using incipient slip
  sensing during a manipulation task,'' in \emph{[1993] Proceedings IEEE
  International Conference on Robotics and Automation}.\hskip 1em plus 0.5em
  minus 0.4em\relax IEEE, 1993, pp. 429--434.

\bibitem{yussof2010sensorization}
H.~Yussof, J.~Wada, and M.~Ohka, ``Sensorization of robotic hand using optical
  three-axis tactile sensor: Evaluation with grasping and twisting motions,''
  \emph{Journal of Computer Science}, 2010.

\bibitem{ito2010shape}
Y.~Ito, Y.~Kim, C.~Nagai, and G.~Obinata, ``Shape sensing by vision-based
  tactile sensor for dexterous handling of robot hands,'' in \emph{2010 IEEE
  International Conference on Automation Science and Engineering}.\hskip 1em
  plus 0.5em minus 0.4em\relax IEEE, 2010, pp. 574--579.

\bibitem{assaf2014seeing}
T.~Assaf, C.~Roke, J.~Rossiter, T.~Pipe, and C.~Melhuish, ``Seeing by touch:
  Evaluation of a soft biologically-inspired artificial fingertip in real-time
  active touch,'' \emph{Sensors}, vol.~14, no.~2, pp. 2561--2577, 2014.

\bibitem{hristu2000performance}
D.~Hristu, N.~Ferrier, and R.~W. Brockett, ``The performance of a
  deformable-membrane tactile sensor: basic results on geometrically-defined
  tasks,'' in \emph{Proceedings 2000 ICRA. Millennium Conference. IEEE
  International Conference on Robotics and Automation. Symposia Proceedings
  (Cat. No. 00CH37065)}, vol.~1.\hskip 1em plus 0.5em minus 0.4em\relax IEEE,
  2000, pp. 508--513.

\bibitem{li2014localization}
R.~Li, R.~Platt, W.~Yuan, A.~ten Pas, N.~Roscup, M.~A. Srinivasan, and
  E.~Adelson, ``Localization and manipulation of small parts using gelsight
  tactile sensing,'' in \emph{2014 IEEE/RSJ International Conference on
  Intelligent Robots and Systems}.\hskip 1em plus 0.5em minus 0.4em\relax IEEE,
  2014, pp. 3988--3993.

\bibitem{yamaguchi2017grasp}
A.~Yamaguchi and C.~G. Atkeson, ``Grasp adaptation control with finger vision:
  Verification with deformable and fragile objects,'' in \emph{the 35th Annual
  Conference of the Robotics Society of Japan (RSJ2017), pp. 1L3--01}, 2017.

\bibitem{yamaguchi2016combining}
------, ``Combining finger vision and optical tactile sensing: Reducing and
  handling errors while cutting vegetables,'' in \emph{2016 IEEE-RAS 16th
  International Conference on Humanoid Robots (Humanoids)}.\hskip 1em plus
  0.5em minus 0.4em\relax IEEE, 2016, pp. 1045--1051.

\bibitem{RV14}
R.~Muthusamy and V.~Kyrki, ``Decentralized approaches for cooperative grasp
  planning,'' in \emph{Int. Conf. Contr. Autom. Robot. Vis. (ICARCV)}, 2014,
  pp. 693--698.

\bibitem{RV2016}
------, ``{Performance metrics for robotic grasping system},'' 2016.

\bibitem{Nguyen1988}
V.~D. Nguyen, ``Constructing force-closure grasps,'' \emph{The Int. J. of
  Robot. Res.}, vol.~7, pp. 3--16, 1988.

\bibitem{takahashi2008adaptive}
T.~Takahashi, T.~Tsuboi, T.~Kishida, Y.~Kawanami, S.~Shimizu, M.~Iribe,
  T.~Fukushima, and M.~Fujita, ``Adaptive grasping by multi fingered hand with
  tactile sensor based on robust force and position control,'' in \emph{2008
  IEEE International Conference on Robotics and Automation}.\hskip 1em plus
  0.5em minus 0.4em\relax IEEE, 2008, pp. 264--271.

\bibitem{gallego2019event}
G.~Gallego, T.~Delbruck, G.~Orchard, C.~Bartolozzi, B.~Taba, A.~Censi,
  S.~Leutenegger, A.~Davison, J.~Conradt, K.~Daniilidis \emph{et~al.},
  ``Event-based vision: A survey,'' \emph{arXiv preprint arXiv:1904.08405},
  2019.

\bibitem{rigi2018novel}
A.~Rigi, F.~Baghaei~Naeini, D.~Makris, and Y.~Zweiri, ``A novel event-based
  incipient slip detection using dynamic active-pixel vision sensor (davis),''
  \emph{Sensors}, vol.~18, no.~2, p. 333, 2018.

\bibitem{naeini2019novel}
F.~B. Naeini, A.~Alali, R.~Al-Husari, A.~Rigi, M.~K. AlSharman, D.~Makris, and
  Y.~Zweiri, ``A novel dynamic-vision-based approach for tactile sensing
  applications,'' \emph{IEEE Transactions on Instrumentation and Measurement},
  2019.

\bibitem{ward2020neurotac}
B.~Ward-Cherrier, N.~Pestell, and N.~F. Lepora, ``Neurotac: A neuromorphic
  optical tactile sensor applied to texture recognition,'' \emph{arXiv preprint
  arXiv:2003.00467}, 2020.

\bibitem{benosman2013event}
R.~Benosman, C.~Clercq, X.~Lagorce, S.-H. Ieng, and C.~Bartolozzi,
  ``Event-based visual flow,'' \emph{IEEE transactions on neural networks and
  learning systems}, vol.~25, no.~2, pp. 407--417, 2013.

\bibitem{clady2015asynchronous}
X.~Clady, S.-H. Ieng, and R.~Benosman, ``Asynchronous event-based corner
  detection and matching,'' \emph{Neural Networks}, vol.~66, pp. 91--106, 2015.

\bibitem{vasco2016fast}
V.~Vasco, A.~Glover, and C.~Bartolozzi, ``Fast event-based harris corner
  detection exploiting the advantages of event-driven cameras,'' in \emph{2016
  IEEE/RSJ International Conference on Intelligent Robots and Systems
  (IROS)}.\hskip 1em plus 0.5em minus 0.4em\relax IEEE, 2016, pp. 4144--4149.

\bibitem{mueggler2017fast}
E.~Mueggler, C.~Bartolozzi, and D.~Scaramuzza, ``Fast event-based corner
  detection,'' in \emph{in 28th British Machine Vision Conference
  (BMVC)}.\hskip 1em plus 0.5em minus 0.4em\relax University of Zurich, 2017.

\bibitem{alzugaray2018asynchronous}
I.~Alzugaray and M.~Chli, ``Asynchronous corner detection and tracking for
  event cameras in real time,'' \emph{IEEE Robotics and Automation Letters},
  vol.~3, no.~4, pp. 3177--3184, 2018.

\bibitem{zhu2017event}
A.~Z. Zhu, N.~Atanasov, and K.~Daniilidis, ``Event-based feature tracking with
  probabilistic data association,'' in \emph{2017 IEEE International Conference
  on Robotics and Automation (ICRA)}.\hskip 1em plus 0.5em minus 0.4em\relax
  IEEE, 2017, pp. 4465--4470.

\bibitem{vidal2018ultimate}
A.~R. Vidal, H.~Rebecq, T.~Horstschaefer, and D.~Scaramuzza, ``Ultimate slam?
  combining events, images, and imu for robust visual slam in hdr and
  high-speed scenarios,'' \emph{IEEE Robotics and Automation Letters}, vol.~3,
  no.~2, pp. 994--1001, 2018.

\end{thebibliography}

\end{document}